\documentclass[sigconf]{acmart}

\usepackage{hyperref}       
\usepackage{url}            
\usepackage{booktabs}       
\usepackage{xcolor}         
\usepackage{mathtools}
\usepackage{amsthm}
\usepackage{bbm}
\usepackage{wrapfig}

\usepackage{multirow}
\usepackage{tabularx}
\usepackage{algorithm}
\usepackage{algorithmic}
\usepackage{subcaption}
\usepackage{enumitem}

\newcolumntype{C}{>{\centering\arraybackslash}X} 
\newtheorem{problem}{Problem}

\newcommand{{\method}}{PreGIP}
\theoremstyle{plain}
\newtheorem{theorem}{Theorem}[section]
\newtheorem{proposition}[theorem]{Proposition}

\newtheorem{corollary}[theorem]{Corollary}
\theoremstyle{definition}
\newtheorem{definition}[theorem]{Definition}

\theoremstyle{remark}

\author{Enyan Dai}
\affiliation{%
  \institution{Hong Kong University of Science and Technology (Guangzhou)}
  \city{Guangzhou}
  \country{China}
}
\email{enyandai@hkust-gz.edu.cn}
\authornote{Both authors contribute equally to this paper.}

\author{Minhua Lin}
\affiliation{%
  \institution{The Pennsylvania State University}
  \city{State College}
  \country{USA}
}
\email{mfl5681@psu.edu}
\authornotemark[1]

\author{Suhang Wang}
\affiliation{%
  \institution{The Pennsylvania State University}
  \city{State College}
  \country{USA}
}
\email{szw494@psu.edu}
\copyrightyear{2025}
\acmYear{2025}
\setcopyright{acmlicensed}\acmConference[KDD '25]{Proceedings of the 31st ACM
SIGKDD Conference on Knowledge Discovery and Data Mining V.2}{August 3--7,
2025}{Toronto, ON, Canada}
\acmBooktitle{Proceedings of the 31st ACM SIGKDD Conference on Knowledge
Discovery and Data Mining V.2 (KDD '25), August 3--7, 2025, Toronto, ON, Canada}
\acmDOI{10.1145/3711896.3737089}
\acmISBN{979-8-4007-1454-2/2025/08}

\begin{document}

\fancyhead{}
\title{PreGIP: Watermarking the Pretraining of  Graph Neural Networks for Deep IP Protection}

\begin{abstract}
Pretraining on Graph Neural Networks (GNNs) has shown great power in facilitating various downstream tasks. As pretraining generally requires huge amount of data and computational resources, the pretrained GNNs are high-value Intellectual Properties (IP) of the legitimate owner. 
However, adversaries may illegally copy and deploy the pretrained GNN models for their  downstream tasks. Though initial efforts have been made to watermark GNN classifiers for IP protection, these methods are not applicable to self-supervised pretraining of GNN models. Hence, in this work, we propose a novel framework named {\method} to watermark the pretraining of GNN encoder for IP protection while maintaining the high-quality of the embedding space. {\method} incorporates a task-free watermarking loss to watermark the embedding space of pretrained GNN encoder. A finetuning-resistant watermark injection is further deployed. Theoretical analysis and extensive experiments show the effectiveness of {\method}. The code can be find in \url{https://anonymous.4open.science/r/PreGIP-semi/} and  \url{https://anonymous.4open.science/r/PreGIP-transfer/}.
\end{abstract}



\maketitle

\section{Introduction}
Graph Neural Networks (GNNs) have shown great power in modeling graphs for various tasks~\cite{xu2024llm,dai2021labeled,dai2021say,dai2022learning}.
Recently, inspired by the success of pretraining on large-scale text data ~\cite{ouyang2022training,lin2024decoding,lin2025far,wu2025lanp}, various self-supervised learning methods such as Graph Contrastive Learning (GCL)~\cite{you2020graph} have been proposed to pretrain GNN models to facilitate the adoption of GNNs on downstream tasks. These pretraining methods have been widely applied for various domains such as molecular property prediction~\cite{hu2020pretraining, lin2025stealing} and protein analysis~\cite{zhang2022protein}, and energy network anomaly detection~\cite{dai2022graph}. 

The pretraining of GNN models generally demands massive computation on large-scale datasets. For example, self-supervised pretraining in chemistry~\cite{hu2020pretraining} takes over 24 hours on an A100 GPU. Similarly, pretraining on proteins~\cite{zhang2022protein,li2025unizyme} is conducted on datasets with around 1M protein structures predicted by AlphaFold2. Consequently, pretrained GNN models are high-value intellectual properties (IP) required to be protected. In practice, the pretrained models are often shared with authorized users for downstream classifier training. However, this poses a significant risk that unauthorized individuals may steal and deploy the pretrained GNN models for their own tasks without the model owner's permission. Therefore, it is crucial to protect the IP of pretrained GNN models.

\begin{figure}
    \centering
    \includegraphics[width=0.92\linewidth]{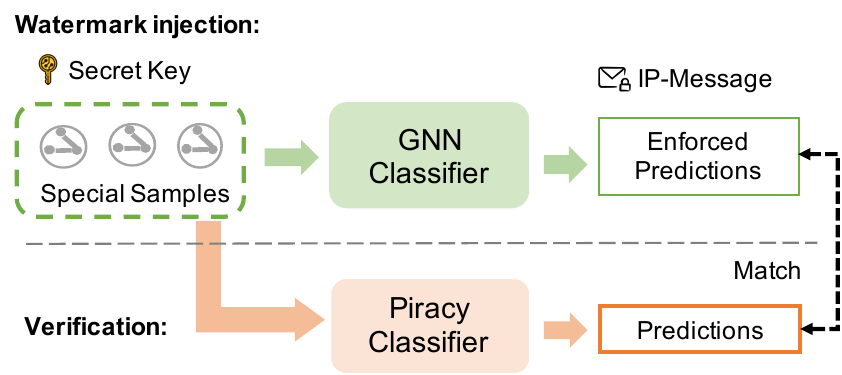}
    \vskip -0.5em
    \caption{GNN classifier watermarking and verification.}
    \vskip -1.5em
    \label{fig:prelimary}
\end{figure}

Several model ownership verification approaches~\cite{adi2018turning,yang2021robust,xu2023watermarking,zhao2021watermarking} have been proposed to protect the IP of GNN classifiers trained in a supervised manner. A typical way of IP protection is to add watermark during the training of GNN classifiers. As illustrated in Fig.~\ref{fig:prelimary}, during the training phase, watermarking approaches will force the protected model to predict trigger samples as certain pre-defined classes~\cite{xu2023watermarking,adi2018turning}. As a result, the predictions on the trigger samples can work as an IP message for model ownership verification. Specifically, a suspect model that gives identical predictions as the protected model on the trigger samples can be identified as a piracy model stolen from the protected model. 
However, the existing watermarking methods are not applicable for pretraining GNNs. This is because pretraining of GNNs only uses pretext tasks to obtain a high-quality embedding space. As downstream tasks and labels are generally unknown at the stage of pretraining, the model owner cannot utilize trigger sample classification in downstream tasks to convey IP messages. 

Therefore, in this paper, we propose a task-free IP protection framework that aims to watermark the representation space for pretraining GNN encoder. Specifically, as shown in Fig.~\ref{fig:framework}, we implement watermarking during pretraining by ensuring similar representations for each pair of dissimilar watermark graphs. If the representations of two paired watermark graphs are close enough, they will receive the same predictions from the downstream classifier under mild conditions, which is verified by our theoretical analysis in Theorem~\ref{theorem:1}. 
By contrast, an independent model trained on non-watermarked data would obtain distinct/dissimilar representations for each watermarked graph pair, resulting in inconsistent predictions. Hence, as shown in Fig.~\ref{fig:framework}, during the verification, we can identify the piracy model by assessing the consistency of its predictions on downstream tasks for each pre-defined watermark graph pair. 

However, watermarking the embedding space during unsupervised pretraining of GNNs to achieve the above goal is a non-trivial task. There are two major challenges to be resolved. \textit{Firstly}, as mentioned above, we aim to ensure similar representations for paired watermark graphs, which will obtain the same predictions from downstream classifiers as the IP message. How can we achieve this and simultaneously preserve the high quality of the embedding space for non-watermark graphs? \textit{Secondly}, to improve the performance on downstream tasks, unauthorized adversaries often finetune the pretrained GNN encoder. The fine-tuning process may erase the watermarks injected into the pretrained GNN model. In an attempt to address the challenges, we propose a novel \underline{Pre}trained \underline{G}NN \underline{IP} protection framework ({\method}). Specifically, {\method} proposes a task-free watermarking loss to guarantee both watermarking performance and discriminability of graph representations. In addition, to reduce the effects of watermarking on real graphs, watermarking graph pairs consist of synthetic graphs. And a finetuning-resistant watermarking approach is deployed in {\method} to ensure the identification of suspect models built from finetuning the protected GNN encoder on various tasks. The effectiveness of the proposed {\method} is demonstrated theoretically and empirically. 
In summary, our main contributions are: 
\begin{itemize}[leftmargin=*]
    \item We study a new problem of watermarking the pretraining of GNN without the downstream task information for deep IP protection.
    \item We propose a novel framework that can watermark the pretraining of GNN and resist the finetuning of the pretrained GNN.
    \item Theoretical analysis is conducted to verify the feasibility of {\method} in watermarking pretraining GNNs.
    \item Extensive experiments on various datasets under different experimental settings show the effectiveness of {\method} in watermarking pretraining GNN encoder and maintaining the high-performance for non-watermark graphs. 
\end{itemize}

\section{Problem Formulation}


\subsection{Preliminaries of Model Watermarking}

Watermarking is an IP protection technique for identifying piracy models. Specifically, a piracy model is an unauthorized copy or finetuned version of a protected AI model, exploited by adversaries for their purposes. 
The model watermarking consists of two stages, i.e., watermark injection and watermark verification. In the watermark injection phase, unique secret keys $\mathcal{K}$ and corresponding IP message $\mathcal{I}$ will be built and injected into the deep model $f$. As Fig.~\ref{fig:prelimary} shows, a typical form of secret keys and IP messages for a supervised GNN classifier are special input samples and corresponding model predictions. When verifying the ownership of model ${f}$, the model owner will extract the IP message with $\mathcal{K}$ by ${\mathcal I} = ExtractIP({f}, \mathcal{K})$. A piracy model will retrain an IP message similar to the protected model, while an independent model will not.

\subsection{Threat Model} 
In this paper, we focus on the IP protection of pretrained GNN encoders. Adversaries will illegitimately deploy the protected GNN encoder $f_E$ to obtain the classifier ${f}_A: \mathcal{G} \rightarrow y$ for their target classification tasks. Parameters of protected GNN encoders may be finetuned by adversaries. The target tasks of adversaries are unknown to the model owner during the watermark injection phase. During the verification phase, we assume a black-box setting, where the model owner can only obtain the predictions of the suspect model on queried samples. This is reasonable because adversaries are typically reluctant to disclose models constructed without authorization.

\subsection{Problem Definition}
In this paper, we focus on pretraining for graph-level tasks. Let $\mathcal{D} = \{\mathcal{G}_i\}_{i=1}^{|\mathcal{D}|}$ denote the set of unlabeled graphs used for GNN encoder pretraining. Given the above description, the problem of watermarking the pretraining of GNNs can be formulated by:

\begin{problem}
    Given a training set $\mathcal{D}$, we aim to construct and inject $(\mathcal{K}, \mathcal{I})$, i.e., secret keys and IP message, into the pretraining of GNN encoder $f_E: \mathcal{G} \rightarrow \mathbb{R}^d$. The watermarked GNN encoder $f_E$ should produce high-quality embeddings effectively in downstream tasks. And when adversaries deploy and finetune watermarked GNN encoder $f_E$ to build unauthorized classifier ${f}_A: \mathcal{G} \rightarrow y$, the IP message can be preserved and extracted from the piracy model ${f}_A$ by $\mathcal{I} = ExtractIP ({f}_A, \mathcal{K})$. ${f}_A$ is composed of a classifier $f_C$ on top of $f_E$. 
\end{problem}

\begin{figure}
    \centering
    \includegraphics[width=0.92\linewidth]{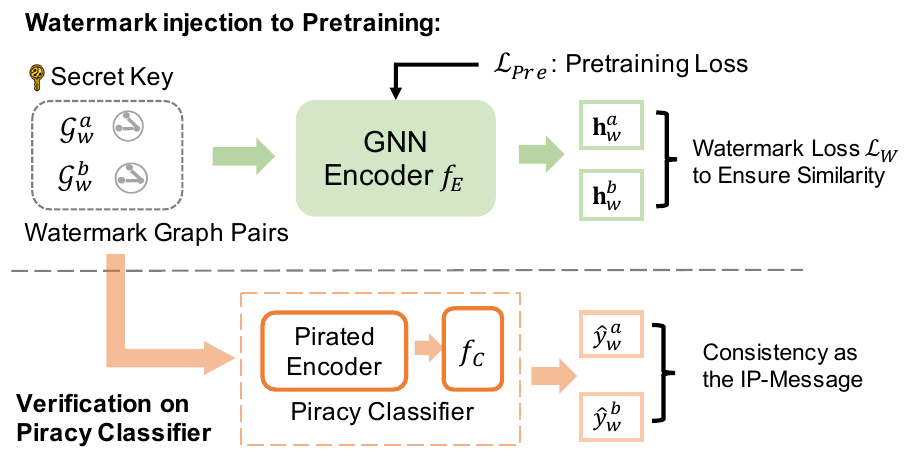}
    \vskip -1em
    \caption{Overview framework of our {\method}.}
    \vskip -1.5em
    \label{fig:framework}
\end{figure}

\section{Methodology}

In this section, we give the details of our proposed {\method}, which is illustrated in Fig.~\ref{fig:framework}. As shown in Fig.~\ref{fig:framework}, {\method} aims to enforce similar representations for multiple pairs of watermark graphs. Then, downstream classifiers topped on the pretrained GNN encoder will give the same predictions on each pre-defined watermark graph pair. As a result, the consistency of predictions on the paired watermark graphs can indicate whether the suspect classifier is constructed from the watermarked GNN encoder.
There are two technical challenges that remain to be addressed: (i) how can we watermark the representation space of the pretrained GNN encoder and simultaneously preserve the high quality of the learned representation space? (ii) adversaries may finetune the pre-trained GNN for various downstream tasks. How to ensure the watermark is preserved after finetuning the pretrained GNN encoder by adversaries? To address the above challenges, {\method} incorporates a task-free watermarking loss that guarantees both watermarking performance and discriminability of graph embeddings. 
A finetuning-resistant watermark injection algorithm is also deployed in {\method}. Next, we give the details of each component.

\subsection{Task-Free Watermarking Framework} \label{sec:task-free}
In this subsection, we give the details of the task-free watermarking framework for IP protection on pretraining GNNs. 

\subsubsection{Overview of Framework} Since adversaries' downstream tasks are generally unavailable for the pretrained GNN owner, we propose a task-free watermarking framework. As presented in Fig.~\ref{fig:framework}, multiple pairs of watermark graphs $\{(\mathcal{G}_w^a, \mathcal{G}_w^b)\}_{w=1}^{|\mathcal{K}|}$ are constructed as secret keys $\mathcal{K}$ for IP protection. During the watermark injection phase, the proposed task-free watermark loss will ensure highly similar representations for each pair of watermark graphs $(\mathcal{G}_w^a, \mathcal{G}_w^b)$, while preserving the quality of representation space.
In Theorem~\ref{theorem:1}, we theoretically show that similar representations will yield the same predictions from the downstream classifier. Thus, the consistency of predictions from the suspect GNN classifier $f_A$ on each pair of $(\mathcal{G}_w^a, \mathcal{G}_w^b)$ can work as an IP message to identify unauthorized deployment of watermarked GNN encoder.
Next, we present the detailed implementation of task-free watermarking.

\subsubsection{Watermark Graph Pair Construction} To watermark pretraining of the GNN encoder, {\method} will enforce similar representations on distinct graphs, leading to identical predictions in the case of the piracy model. One straightforward way of construction is to randomly sample real graphs as watermark graph pairs. However, enforcing close embeddings for distinct real graphs may significantly degrade the embedding quality on real-world test samples. Therefore, we utilize synthetic graphs to construct the watermark graph pair set $\mathcal{K} = \{(\mathcal{G}_w^a, \mathcal{G}_w^b)\}_{w=1}^{|\mathcal{K}|}$. Specifically, the node attributes $\mathbf{x}$ of each graph in $\mathcal{K}$ are sampled by:
\begin{equation}
    \mathbf{x} \sim \mathcal{N}(\bf{\mu}, \bf{\sigma}^2),
    \label{eq:node_sample}
\end{equation}
where $\mu$ and $\sigma$ denote the mean and standard deviation of the node attributes in the pretraining dataset $\mathcal{D}$, respectively. 
As for the graph structure generation, we use an Erdos-Renyi (ER) random graph model~\cite{erdds1959random} which independently connects each pair of nodes with probability $p$. Note that for each pair of watermark graphs  $(\mathcal{G}_w^a, \mathcal{G}_w^b)$, their node features and graph structures are independently synthesized, resulting in two distinct graphs. 
Additionally, the probability parameter $p$ and the size of the graph will be set as different values to further make $\mathcal{G}_w^a$ and $\mathcal{G}_w^b$ distinct. In this way, $\mathcal{G}_w^a$ and $\mathcal{G}_w^b$ are ensured to receive dissimilar predictions from a model independently trained on the non-watermarked dataset.

\subsubsection{Watermarking Loss}
Apart from the term for high similarity between paired watermark graphs, we also incorporate a loss term to encourage that embeddings of synthetic watermark graphs have no overlap with those of real-world graphs. This will offer two advantages over the loss that solely enforces similar embeddings between watermark graph pairs: (i) it prevents the trivial solution that $f_E$ outputs the same embeddings to all graphs; (ii) embedding space of synthetic watermark graphs will remain distinct from that of real-world graphs, thereby preserving the quality of real graph embeddings for downstream tasks. 

Specifically, let $\mathbf{h}_w^a$ and $\mathbf{h}_w^b$ denote the graph representations of each pair of watermark graphs $(\mathcal{G}_w^a, \mathcal{G}_w^b)$ generated by $f_E$, the watermarking loss is formulated as:
\begin{equation} 
\label{eq:watermark}
\small
\begin{aligned}
         \min_{\theta} & \quad  \mathcal{L}_{W}(\theta)  =  \sum_{(\mathcal{G}_w^a, \mathcal{G}_w^b) \in \mathcal{K}} \Big( \| \mathbf{h}_w^a- \mathbf{h}_w^b \|^2 + \\ 
         &\sum_{\mathcal{G}_w^k \in (\mathcal{G}_w^a, \mathcal{G}_w^b)} \sum_{i=1}^Q \mathbb{E}_{\mathcal{G}_i \sim \mathcal{D}} \max (0, m-\|\mathbf{h}_w^k- \mathbf{h}_i\|^2) \Big),
\end{aligned}
\end{equation}
where $\theta$ represents parameters of GNN encoder $f_E$ and $\mathcal{G}_i$ will be randomly sampled from the pretraining set $\mathcal{D}$. $Q$ is the number of samples. The latter term in Eq.(\ref{eq:watermark}) will ensure the margin between the synthetic watermark graphs and real-world graphs is maintained at a minimum of $m$ in the representation space. With Eq.(\ref{eq:watermark}), the whole process of pretraining with  watermark embedding can be written as:
\begin{equation} \label{eq:wm_with_pre} 
    \min_{\theta} \quad \mathcal{L}_{Pre}(\theta) + \lambda \mathcal{L}_W(\theta),
\end{equation}
where $\mathcal{L}_{Pre}$ denotes the pretraining loss such as InfoNEC loss~\cite{oord2018representation} and edge prediction loss~\cite{hu2020pretraining}.  $\lambda$ is the hyperparameter to balance the utility and watermarking.

\subsubsection{Ownership Verification} For a piracy classifier that utilizes the watermarked GNN encoder $f_E$, it will be more likely to give identical predictions on each pair of $(\mathcal{G}_w^a, \mathcal{G}_w^b)$; while an independently trained classifier would not exhibit this behavior. Thus, we extract the IP message by computing the portion of the same predictions on the watermark graph pair set $\{(\mathcal{G}_w^a, \mathcal{G}_w^b)\}_{w=1}^{|\mathcal{K}|}$. Given a GNN classifier $f_A: \mathcal{G} \rightarrow y$ to be verified, its IP indication score for ownership verification is:
\begin{equation} \label{eq:IPScore}
    I(f_A) = \frac{\sum_{w=1}^{|\mathcal{K}|} \mathbbm{1} (f_A(\mathcal{G}_w^a)=f_A(\mathcal{G}_w^b))}{|\mathcal{K}|},
\end{equation}
where $\mathbbm{1}(\cdot)$ denotes the indicator function that returns 1 if predictions on paired watermark graphs from $f_A$ are matched. IP indication score $I(f_A)$ ranges from 0 to 1. A high $I(f_A)$ indicates the classifier $f_A$ is pirated from the watermarked GNN encoder $f_E$.

\subsubsection{Theoretical Analysis}
We conduct the following theoretical analysis to justify the feasibility of the proposed task-free watermarking framework. 
\begin{theorem} \label{theorem:1}
    We consider a $K$-layer MLP downstream classifier $f_C: \mathbb{R}^d \rightarrow \mathbb{R}^{c}$ with 1-Lipschitz activation function (e.g., ReLU, sigmoid, tanh) that inputs graph embeddings learned by pretrained GNN encoder $f_E: \mathcal{G} \rightarrow \mathbb{R}^d$. Let $s_a$ denote the margin between the logit score (unnormalized score before softmax) of the predicted class and the second confident class by $f_C$ on a watermark graph $\mathcal{G}_w^a$.
    The predicted class on the paired watermark graph $\mathcal{G}_w^b$ from $f_C$ is guaranteed to be the same as that of $\mathcal{G}_w^a$ when:
    \begin{equation} 
    \small
        \|f_E(\mathcal{G}_w^a) - f_E(\mathcal{G}_w^b) \|_2 < \frac{1}{2} \cdot \frac{s_a}{\prod_{i=1}^K \|\mathbf{W}_i\|_2},
    \end{equation}
    where $\mathbf{W}_i$ denotes the parameters of $i$-th layer in the downstream classifier $f_C$.  
\end{theorem}
Please refer to the Appendix~\ref{app:proof1} for detailed proof. According to Theorem~\ref{theorem:1}, the downstream classifier $f_C$ will predict the same label for $\mathcal{G}_w^a$ and $\mathcal{G}_w^b$ as long as their representations obtained by the watermarked encoder $f_E$ are close enough. Additionally, the potential adoption of the $l_2$-norm regularization in downstream classifier training will largely constrain the value of $\|\mathbf{W}_i\|_2$,  which can ease the fulfillment of the condition specified in Theorem~\ref{theorem:1}.

\subsection{Finetuning-Resistant Watermark Injection}
In this subsection, we first analyze how the finetuning on the watermarked GNN encoder will affect the injected watermarks. Then, we introduce the finetuning-resistant watermark injection. 

\subsubsection{Impacts of Finetuning to Watermarking} To further improve the prediction accuracy on target tasks, adversaries often finetune the stolen pretrained GNN encoder. The magnitude of modifications applied to the GNN encoder is usually limited, given that the pretrained GNN encoder already offers an effective initialization.  The directions of the GNN parameter updating in finetuning can be diverse, as adversaries may apply the pretrained GNN for various tasks. Therefore, following~\cite{bansal2022certifiedwm}, we suppose the pretrained GNN parameters $\theta$ are updated within a spherical radius $\epsilon$ in our analysis, i.e., $\|\hat{\theta}-\theta\|_2 \leq \epsilon$, where $\hat{\theta}$ denotes the updated parameters of the GNN encoder $f_E$ and $\epsilon$ should be a relatively small value. Let $s_a$ denote the margin between logit scores of the predicted class and the second confident class on a watermark graph $\mathcal{G}_w^a$ by the downstream classifier. Following~\cite{cohen2019certified}, we also assume the lower bound of $s_a$, denoted as $\underline{s}_a$ is larger than 0, which is formaluted as $s_a \geq \underline{s_a}>0$.
With the above assumptions and Theorem~\ref{theorem:1}, we have the following corollary.
\begin{corollary} \label{theorem:2} 
     We consider a $K$-layer MLP downstream classifier $f_C: \mathbb{R}^d \rightarrow \mathbb{R}^{c}$ with 1-Lipschitz activation function is topped on the  $f_E$ finetuned by adversaries.
    Suppose (i) the parameters of pretrained GNN encoder $f_E$ are finetuned to $\hat{\theta}$, where $\|\hat{\theta}-\theta\|_2 \leq \epsilon$; (ii) $s_a \geq \underline{s_a} > 0$, where $s_a$ denotes the margin between logit score (unnormalized score before softmax) of the predicted class and the second confident class by $f_C$ on a watermark graph $\mathcal{G}_w^a$.  
    The predicted class on the paired watermark graph $\mathcal{G}_w^b$ from $f_C$ is guaranteed to the same as that of $\mathcal{G}_w^a$ when:
    \begin{equation} \small
       \sup_{\| \delta \| \leq \epsilon} \|f_E(\mathcal{G}_w^a; \theta + \delta) - f_E(\mathcal{G}_w^b; \theta+\delta) \|_2 < \frac{1}{2} \cdot \frac{\underline{s_a}}{\prod_{i=1}^K \|\mathbf{W}_i\|_2},
    \end{equation}
    where $\mathbf{W}_i$ denotes the parameters of $i$-th layer in the downstream classifier $f_C$.  
\end{corollary}
The steps of deriving Corollary~\ref{theorem:2} are in Appendix~\ref{app:proof2}. 

\subsubsection{Finetuning-Resistant Watermarking}  
According to Corollary~\ref{theorem:2}, to ensure the effectiveness of watermarking after fine-tuning for downstream tasks, we need to achieve low watermarking loss for any update $\delta, \|\delta\|_2 \leq \epsilon$, on the model parameters $\theta$. Simultaneously, the GNN encoder parameters $\theta$ should maintain low loss for the pretraining pretext tasks to preserve the utility. To achieve the above objectives, we replace the objective function of watermark embedding in Eq.(\ref{eq:wm_with_pre}) with  the following objective function to optimize parameters $\theta$ of the GNN encoder $f_E$:
\begin{equation}
\label{eq:FRWater}
\begin{aligned} \small
    & \min_{\theta} \mathcal{L}_{Pre}(\mathcal{D}, \theta) + \lambda \mathcal{L}_{W}(\theta + \delta^*) \\
    &\text{s.t.} \quad \delta^* = \arg \max_{\|\delta\| \leq \epsilon} \mathcal{L}_{W}(\theta + \delta^*),
\end{aligned}
\end{equation}
where $\mathcal{L}_{Pre}$ denotes the pretraining loss and $\lambda$ is the hyperparameter to balance the utility and watermarking. $\epsilon$ can be tuned for different magnitudes of potential modifications.  
With the Eq.(\ref{eq:FRWater}), we will obtain a watermarked GNN encoder that preserves the watermarks after its model parameter $\theta$ is updated within the $\epsilon$-ball. And pretraining loss and the specially designed watermarking loss also guarantee the utility of the watermarked GNN encoder in downstream tasks.  
\begin{table*}[t]
    \centering
    \small
    \caption{Results in protecting IP of GCL-pretrained GNNs under the semi-supervised learning setting.}
    \vskip -1em
    {\begin{tabular}{lllcccccccc}
    \toprule
    Scenario & {Dataset}  & 
    {Metrics (\%)} & Non-Watermarked  & WPE & SSL-WM & {GCBA}& {Deepsign}& {RWM-Pre}& {CWM-Pre}& {PreGIP}\\
     \midrule
    \multirow{9}{*}{Fix} &  
    
    \multirow{3}{*}{PROTEINS} & {{Accuracy} } & 70.5$\pm$1.7 & 69.7$\pm$1.6 &  59.7$\pm$2.0 &  70.3$\pm$1.5 & 69.2$\pm$1.5 & 70.1$\pm$1.7 & 69.2$\pm$1.7 & \textbf{70.3$\pm$1.1}\\
    &  & {{IP Gap} } & - & 18.4$\pm$16.2 & 23.6$\pm$3.0 & 3.0$\pm$1.0 & 11.9$\pm$0.5 & 16.4$\pm$1.3 & 32.4$\pm$1.1 & \textbf{46.5$\pm$2.2}\\
    &  & IP ROC & -  & 79.1 & \textbf{100} & 80.0& 85.1& 87.7& 93.4& \textbf{100}\\

    \cmidrule{2-11}
    & \multirow{3}{*}{NCI1}
    &  {{Accuracy} } & 66.2$\pm$1.1 & 64.1 $\pm$1.9 & 63.3 $\pm$3.3 &  65.8$\pm$0.4 & 65.0$\pm$0.8 & 65.7$\pm$1.1 &  65.8$\pm$0.9 & \textbf{65.9$\pm$0.4}\\
    & & {{IP Gap} } & - & 14.5$\pm$5.6 & 2.3$\pm$6.2 & 2.2$\pm$1.6 & 1.6$\pm$0.2 & 16.6$\pm$7.9 & 6.0$\pm$3.9 & \textbf{60.0$\pm$4.9}\\
    & & IP ROC  & - & 62.1 & 69.1  & 56.2 & 59.8 & 85.5 & 64.8 & \textbf{100}\\
    
    \cmidrule{2-11}
    & \multirow{3}{*}{FRANKS.}
    &  {{Accuracy} } & 60.8$\pm$1.3 & 59.5$\pm$1.1 & 60.3$\pm$0.9 & 59.6$\pm$0.7 & 60.5$\pm$1.2 & 60.0$\pm$1.0 & 59.9$\pm$1.1 & \textbf{60.5$\pm$0.9}\\
    &  & {{IP Gap} } & - & 25.9$\pm$20.8 & 4.5$\pm$1.5 & 4.0$\pm$2.8 & 13.6$\pm$1.6 & 16.0$\pm$2.7 & 5.2$\pm$3.4 & \textbf{79.5$\pm$24.2}\\
    &  &  IP ROC  & - & 69.8 & 40.3 & 65.9 & 91.4 & 81.9 & 49.8 & \textbf{100}\\

    \midrule
    \multirow{9}{*}{Finetune} & \multirow{3}{*}{PROTEINS}
    & {{Accuracy} } & 70.3$\pm$1.8 & 64.6$\pm$1.2 & 64.7$\pm$1.2 &  67.5$\pm$1.8 & 69.2$\pm$1.5 & 69.3$\pm$2.0 & \textbf{69.8$\pm$1.7} & {69.6$\pm$0.4}\\
    & & {{IP Gap} } & - & 15.3$\pm$8.8 & 2.6$\pm$2.7 & 7.3$\pm$2.9 & 18.4$\pm$1.7 & 6.6$\pm$5.7 & 13.0$\pm$3.7 & \textbf{55.5$\pm$13.5}\\
    & & IP ROC & - & 88.8 & 68.2 & 94.0& 85.6& 58.5& 71.4& \textbf{94.5}\\ 
    \cmidrule{2-11}
    & \multirow{3}{*}{NCI1}
    & {{Accuracy} } & 67.6$\pm$1.0& 67.5$\pm$1.3 & 67.7$\pm$1.3 & 64.3$\pm$1.1 & 67.3$\pm$1.4 & 67.9$\pm$1.3 &  68.0$\pm$1.2 & \textbf{68.8$\pm$0.9}\\
    & & {{IP Gap} } & - & 8.5$\pm$7.4 & 8.4$\pm$5.6 & 16.2$\pm$3.0  & 14.8$\pm$7.2 & 1.3$\pm$0.6 & 4.1$\pm$2.4 & \textbf{28.5$\pm$0.8}\\
    & & IP ROC  & - & 51.8 & 52.6 & 68.2 & 81.2 & 52.5 & 59.9 & \textbf{99.0}\\ 
    \cmidrule{2-11}
    & \multirow{3}{*}{FRANKS.}
    & {{Accuracy} } & 61.7$\pm$1.0 & 59.6$\pm$1.6 & 61.5$\pm$1.4 & 59.3$\pm$1.5 & 60.7$\pm$1.3 & 61.1$\pm$1.5 & 58.6$\pm$1.2 & \textbf{61.1$\pm$1.5}\\ 
    &  & {{IP Gap} } & - & 25.2$\pm$21.1 & 26.3$\pm$25.9 &  3.2$\pm$1.9 & 6.7$\pm$3.8 & 10.7$\pm$9.5 & 21.3$\pm$13.6 & \textbf{27.5$\pm$4.3}\\
    &  & IP ROC & - & 61.8 & 83.0 & 55.5 & 58.0 & 68.3 & 56.2 & \textbf{89.5}\\
    \bottomrule 
    \end{tabular}}
    \label{tab:semi_GraphCL_IP_performance}
\end{table*}

\begin{algorithm}[t] 
\caption{Training Algorithm of {\method}} 
\begin{algorithmic}[1]
\REQUIRE Pretraining dataset $\mathcal{D}$, pretraining method, $\lambda$, $\epsilon$
\ENSURE Watermarked pretrained GNN encoder $f_E$, and watermark graph pairs $\mathcal{K}$ as secret keys 

\STATE Construct a set of watermark graph pairs $\mathcal{K}$  
\STATE Randomly initialize $\theta$ for $f_E$
\WHILE{not converged}
    \FOR{t= $1,2,\dots,T$}
        \STATE Update the $\delta$ with gradient ascent on the watermarking loss $\mathcal{L}_W(\theta+\delta)$ with Eq.(\ref{eq:inner})
    \ENDFOR
    \STATE Calculate the pretraining loss $\mathcal{L}_{Pre}(\theta)$
    \STATE Obtain the gradients for GNN encoder $f_E$ by Eq.(\ref{eq:outer_appr})
    \STATE Update the GNN encoder $f_E$ with gradient decent
\ENDWHILE
\end{algorithmic}
\label{alg:algorithm}
\end{algorithm}
\subsection{Optimization Algorithm}
To efficiently solve the optimization problem in Eq.(\ref{eq:FRWater}), we propose the following alternating optimization schema.

\noindent \textbf{Inner Loop Optimization}. In the inner loop, we update $\delta$ with $T$ iterations of gradient ascent within the $\epsilon$-ball to approximate $\delta^*$:
\begin{equation}
\label{eq:inner} 
    \delta_{t+1} = \delta_t + \alpha \nabla_{\delta} \mathcal{L}_W(\theta+\delta_t),
\end{equation}
where the step size $\alpha$ is set as $\alpha = \epsilon / (T \cdot \max(\|\nabla_{\delta} \mathcal{L}_W(\theta+\delta_t)\|_2,1))$ for optimization under the constraint $\|\delta\|_2\leq \epsilon$.

\noindent \textbf{Outer Loop Optimization on Encoder}. In the outer loop optimization, we need to compute the gradients by:
\begin{equation}
\label{eq:outer}
\begin{aligned} \small
    \nabla_{\theta}^{outer} &  = \nabla_{\theta} \mathcal{L}_{Pre}(\mathcal{D}, \theta) + \lambda \nabla_{\theta} \mathcal{L}_W(\theta+\delta_T).
\end{aligned}
\end{equation}
However, the latter term in Eq.(\ref{eq:outer}) is expensive to compute as $\delta_T$ is a function of $\theta$. We need to unroll the training procedure of $\delta_T$ to calculate gradients. Following~\cite{nichol2018first}, we approximate the gradient by:
\begin{equation} \label{eq:outer_appr} \small
    \nabla_{\theta}^{outer}  = \nabla_{\theta} \mathcal{L}_{Pre}(\mathcal{D}, \theta) + \lambda {\sum}_{t=1}^T \nabla_{\theta} \mathcal{L}_W(\theta+\bar{\delta}_t),
\end{equation}
where $\bar{\delta}_t$ indicates gradient propagation stopping. With Eq.(\ref{eq:outer_appr}), gradient descent can be applied to update encoder parameters $\theta$.  

The training details are given in Algorithm~\ref{alg:algorithm}. In lines 1-2, we synthesize watermark graph pairs and initialize the model parameters.  The inner loop optimization is conducted in lines 4-6. Codes in lines 7-9 correspond to the outer loop optimization.

\subsection{Time Complexity Analysis}
\label{sec:timecomplexity}
In this section, we analyze the time complexity of the finetuning-resistant watermarking injection. 
For the watermarking loss, the major cost comes from computing the representations of watermarked graphs and sampled graphs in batch training. In addition, the watermarking loss will be computed for $T$ steps in each iteration of finetuning-resistant watermarking. Therefore, the overall time complexity of {\method} in watermarking is $O(hT|\mathcal{K}|\cdot (| \mathcal{E}_w | + Q)) $, where $h$ is the hidden dimension of GNN encoder, $Q$ is the batch size of training, and $T$ is the number of inner iteration in optimization. $|\mathcal{K}| $ and $|\mathcal{E}_w|$ denote the number of watermark graphs and edges in each watermark graph. The number of inner iteration in optimization $T$ is generally set as 3 in PreGIP. Furthermore, the number of watermark graphs are also set to a small number such as 20. 
Note that the time complexity of contrastive loss is $O(h(Q+1)^2 \cdot |\mathcal{E}|)$, where $|\mathcal{E}|$ denotes the average number of edge in each pretraining graph. As a result, the additional computation cost from watermarking is small compared to the pretraining loss.


\begin{table*}[t]
    \centering
    \small
    \caption{Results in protecting IP of GCL-pretrained GNNs in the transfer learning setting.}
    \vskip -1em
    {\begin{tabularx}{0.98\linewidth}{XX>{\centering\arraybackslash}p{0.13\linewidth}CCCCCCC}
        \toprule 
        {Dataset} & 
        {Metrics (\%)} & Non-Watermarked & WPE & SSL-WM & {Deepsign}& {RWM-Pre}& {CWM-Pre}& {PreGIP}\\
        \midrule
        \multirow{3}{*}{Tox21} 
        & Accuracy & 92.8$\pm$0.2 & 92.7$\pm$0.2 & 92.2$\pm$0.1 & 92.2$\pm$0.1 & 92.8$\pm$0.2 & 92.7$\pm$0.2 & \textbf{92.8$\pm$0.1}\\
            
        & {{IP Gap} } & - & 2.0$\pm$0.2 & 0.3$\pm$0.1 & 0.0$\pm$0.0 & \textbf{14.3$\pm$9.7}  & 8.1$\pm$10.2 & 8.1$\pm$7.1\\
        &  IP ROC & - & 60.0 & 60.0 & 50.0& 56.0& 74.0& \textbf{80.0}\\
        \midrule
        \multirow{3}{*}{ToxCast}
        &  {{Accuracy} } & 83.4$\pm$0.2 & 83.1$\pm$0.6  & 83.1$\pm$0.3 & 83.2$\pm$0.2 & 83.5$\pm$0.2 &  83.3$\pm$0.1 & \textbf{83.5$\pm$0.2}\\
        & IP Gap& - & 6.4$\pm$3.9 & 4.1$\pm$2.1 & 1.1$\pm$0.4 & 4.7$\pm$4.0 & 3.6$\pm$0.7 & \textbf{7.3$\pm$3.1}\\
        & IP ROC & - & 60.0& 60.0 & 95.0 & 64.0 & 84.0 & \textbf{100}\\
        \midrule
        \multirow{3}{*}{BBBP}
        & {{Accuracy} } & 84.9$\pm$0.8 & \textbf{84.9$\pm$1.2} & 81.0$\pm$1.2 & 81.9$\pm$2.2 & 84.7$\pm$0.9 &  84.3$\pm$1.3 & {84.5$\pm$0.6}\\ 
            
        & {{IP Gap} } & - & 47.5$\pm$30.1  & 6.6$\pm$4.5 & 0.0$\pm$0.0 & \textbf{51.0$\pm$36.9} & 25.0$\pm$11.4 & 49.0$\pm$4.1\\
        & IP ROC & - & 80.0 & 80.0 & 50.0 & 95.0 & 88.0 & \textbf{100}\\
        \midrule
        \multirow{3}{*}{BACE}
        & {{Accuracy} } & 73.4$\pm$0.9 & 74.0$\pm$1.3 & 63.4$\pm$0.3  & 71.2$\pm$2.1 & 74.3$\pm$2.9 &  \textbf{74.9$\pm$1.6} & 74.3$\pm$2.1\\
        & {{IP Gap} } & - & 16.3$\pm$8.5 & 4.5$\pm$2.5 & 7.0$\pm$4.0 & 16.8$\pm$12.3 & 14.5$\pm$10.0 & \textbf{23.0$\pm$18.8}\\
        &  IP ROC & - & 80.0 & 60.0 & 80.0 & 56.0 & 64.0 & \textbf{92.0}\\
        
        \midrule
        \multirow{3}{*}{SIDER}
        & {{Accuracy} } & 75.0$\pm$0.8 & 74.5$\pm$0.6 & 74.9$\pm$0.3 & 74.8$\pm$0.2 & 75.0$\pm$0.4 &  74.8$\pm$0.2 & \textbf{75.0$\pm$0.4}\\ 
            
        & {{IP Gap} } & - & 0.2$\pm$0.1 & 0.0$\pm$0.0 & 1.4$\pm$0.6 & 2.1$\pm$1.6 & 4.3$\pm$3.3 & \textbf{18.3$\pm$10.5}\\
        & IP ROC & - & 70.0 & 50.0 & 68.0 & 52.0 & 64.0 & \textbf{88.0}\\
        \midrule
        \multirow{3}{*}{ClinTox}
        & {{Accuracy} } & 92.3$\pm$1.2 & 92.1$\pm$1.1 & 92.1$\pm$1.3 & 91.9$\pm$1.3 & 91.7$\pm$1.0 &  92.1$\pm$1.3 & \textbf{92.2$\pm$1.3}\\
        & {{IP Gap} } & - & 1.3$\pm$1.1 & 0.0$\pm$0.0 & 0.0$\pm$0.0 & 21.0$\pm$7.0 & 11.5$\pm$13.2 & \textbf{37.5$\pm$43.4}\\
        &  IP ROC & - & 60.0 & 50.0& 50.0 & 52.0 & 80.0 & \textbf{80.0}\\
        \bottomrule 
    \end{tabularx}}
    \label{tab:transfer_GraphCL_IP_performance}
\end{table*}

\section{Experiments}
In this section, we conduct experiments to answer the following research questions.
\begin{itemize}[leftmargin=*]
    \item \textbf{Q1:} Can our {\method} accurately identify piracy models that utilize the watermarked pretrained GNN for their downstream classification tasks under various scenarios?
    \item \textbf{Q2:} Can the watermark injected by our {\method} maintain valid under various watermark removal methods?   
    \item \textbf{Q3:} How do the proposed watermarking loss and the finetuning-resistant watermarking affect the pretrained model in embedding quality and IP protection?
\end{itemize}
\subsection{Experimental Setup}
\label{sec:experimental_setup}
\subsubsection{Datasets} We conduct experiments on widely used benchmark datasets for semi-supervised~\cite{you2020graph} and transfer learning~\cite{hu2020pretraining} on graph classification. Specifically, for the semi-supervised setting, we use three public benchmarks, i.e., PROTEINS, NCI1, and FRANKENSTEIN~\cite{sergey2020understanding}. For the transfer learning setting, we use 200K unlabeled molecules sampled from the ZINC15 dataset~\cite{sterling2015zinc} for pretraining, and use 6 public molecular graph classification benchmarks~\cite{wu2018moleculenet} for the downstream tasks. More details of these datasets and learning settings are in Appendix~\ref{appendix:dataset_statistics}. 

\subsubsection{Baselines}
As the methods in watermarking pretraining GNN encoder are rather limited, we propose several baselines adapted from existing watermarking methods. Firstly, we extend \textbf{WPE}~\cite{wu2022watermarking} and \textbf{SSL-WM}~\cite{lv2022ssl}, which are start-of-the-art methods for watermarking pretraining image encoder.
We also compare with \textbf{DeepSign}~\cite{rouhani2018deepsigns} which embeds watermarks into the hidden-layer activations. The verification phase of DeepSign is adjusted for black-box verification. A state-of-the-art graph model watermarking approach~\cite{zhao2021watermarking} is also compared.  
This method was originally proposed for watermarking supervised classifiers which enforces random graphs to be predicted to a certain class. 
We extend it to a version named \textbf{RWM-Pre} for unsupervised encoder watermarking. To show the effectiveness of our finetuning-resistant watermark injection, we further apply the unmovable watermarking technique~\cite{bansal2022certifiedwm} to RWM-Pre, resulting in a baseline \textbf{CWM-Pre}.
As a backdoor is a widely used solution for watermarking, we further compare a graph contrastive learning backdoor attack method \textbf{GCBA}~\cite{zhang2023graph}.
More details on these baselines are given in Appendix~\ref{appendix:baselines}. 

\subsubsection{Implementation Details} We conduct experiments in watermarking GNN models pretrained by GraphCL~\cite{you2020graph} and edge prediction~\cite{hu2020pretraining}, which are representative works in contrastive pretraining and generative pretraining, respectively. GraphCL is implemented based on PyGCL library~\cite{zhu2021pygcl}. The edge prediction and edge masking strategies are implemented based on the source code published by the original authors. Following GraphCL~\cite{you2020graph}, a 2-layer GIN is employed as the GNN encoder. The hidden dimension is set as 256 for all experiments.  Then, following~\cite{hu2020pretraining}, we employ the obtained embeddings to train and evaluate a classifier for the downstream tasks.   We configure the number of watermark graph pairs as $20$ and set $\lambda=0.1$, $\epsilon=2$ for all experiments. The hyperparameter analysis is given in Sec.~\ref{sec:hyper}.
In the watermark graph pair generation, the probability of connecting each pair of nodes is set as 0.2. We maintain a consistent difference of 15 nodes in the node count between watermark graphs within each pair. By leveraging the randomness in synthetic structures and varying the sizes of the graphs, we are able to generate distinct synthetic graphs to build multiple pairs watermark graphs.
In the watermarking loss, the $Q$ and $m$ in Eq.(\ref{eq:watermark}) are set as 256 and 1.0. 
To make a fair comparison, the size of watermark graphs for baselines is also set as 20. Additionally, all the hyperparameters of these methods are tuned based on the watermarking loss and the pretraining loss on validation set.  All models undergo training on an A6000 GPU with 48G memory. We use Intel Xeon W-3323 CPU.

\subsubsection{Evaluation Metrics}
\textit{Firstly}, we evaluate the quality of the embeddings produced by the watermarked GNN encoder. The pretraining with watermark injection is conducted on the pretraining dataset. Then, with the training set for the downstream task, we train linear classifiers that on top of the pretrained GNN encoder. The \textbf{accuracy} of the linear classifier on the test set for the downstream task is used to assess embeddings generated by the watermarked GNN encoder. We compute the average accuracy across 50 different train/test splits on the downstream task.
\textit{Secondly},  we evaluate whether the watermarking methods can distinguish the piracy models from independent models. In the evaluation,  a piracy model $f_A$ is built by adopting watermarked pretrained GNN in downstream task classification. We obtain multiple piracy models by varying splits in downstream tasks.  An independent classifier $f_G$ will use an independently pretrained GNN that is not watermarked. By varying dataset splits and random seeds in pretraining, we obtain multiple independent models. We adopt the following metrics to evaluate the watermarking performance, i.e., whether one can distinguish the
piracy models from independent models: 
\begin{itemize}[leftmargin=*]
    \setlength{\itemsep}{0pt}
    \item \textbf{IP Gap} calculates the gap between the piracy model $f_A$ and the independent GNN $f_G$ in IP indication score by $\text{IP Gap} = |I(f_A)-I(f_G)|$, where $I(\cdot) \in [0,1]$ is the IP indication score of the watermarking method to infer whether the test model is copied from the protected model. The computation of $I(\cdot)$ for PreGIP and baselines is given in Eq.(\ref{eq:IPScore}) and Appendix~\ref{app:implementation}. A larger IP Gap indicates a stronger watermarking method. 
    \item \textbf{IP ROC} evaluates how well the IP indication score $I$ can separate the piracy classifiers and independent classifiers. Specifically, the built piracy classifiers and independent classifiers are viewed as positive samples and negative samples, respectively. Then, the ROC score is applied based on the IP indication score. 
\end{itemize}

\begin{figure}[t]
    \small
    \centering
    \begin{subfigure}{0.49\linewidth}
        \includegraphics[width=0.98\linewidth]{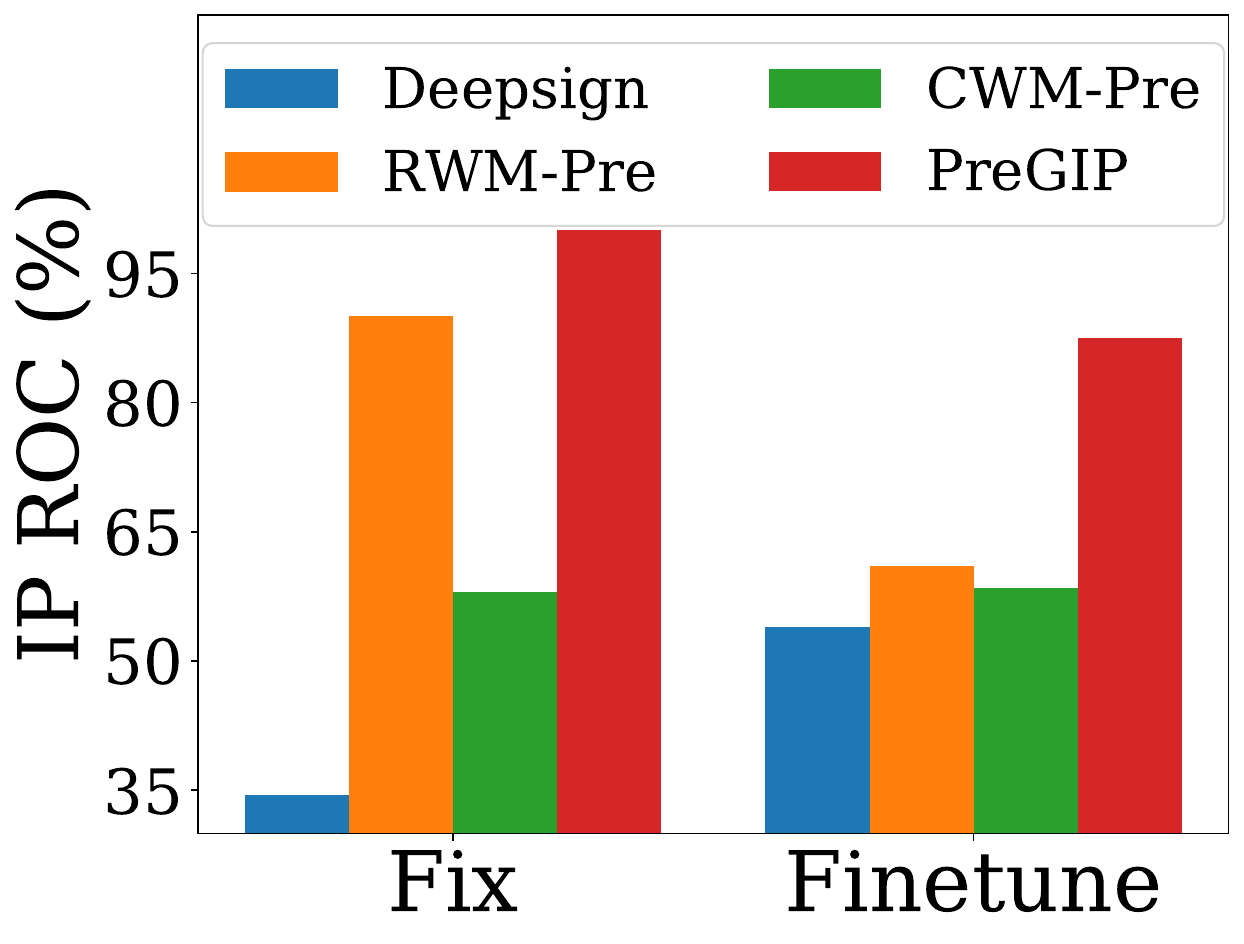}
        \vskip -0.5em
        \caption{IP ROC}
    \end{subfigure}
    \begin{subfigure}{0.49\linewidth}
        \includegraphics[width=0.98\linewidth]{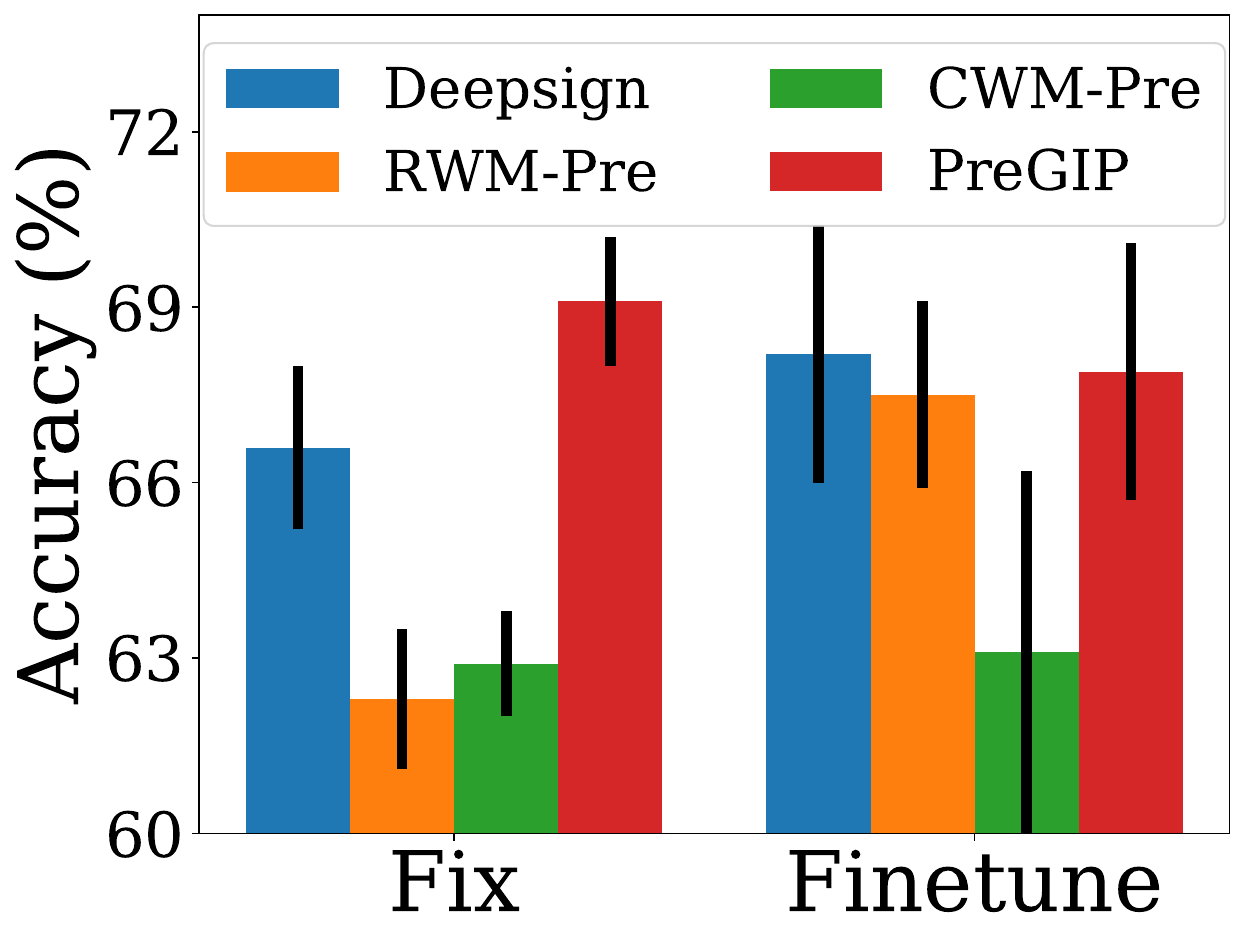}
        \vskip -0.5em
        \caption{Accuracy}
    \end{subfigure}
    \vskip -1 em
    \caption{Watermarking edge prediction-pretrained GNN.}
    \vskip -1 em
    \label{fig:flexibility_of_watermarking}
\end{figure}

\subsection{Results of Watermarking Pretraining}
\label{sec:results_of_watermarking_pretraining}
To answer \textbf{Q1}, we first watermark the contrastive learning for two different pretraining settings, i.e., semi-supervised learning and transfer learning. Then, we conduct experiments in watermarking generative pretraining to show the flexibility of PreGIP in watermarking various GNN pretraining methods. 

\vskip 0.3em
\noindent \textbf{Watermarking Pretraining for Semi-Supervised Learning}.
Following~\cite{you2020graph}, a scheme of semi-supervised learning for downstream graph classification is implemented. Particularly, the GNN is firstly pretrained on the whole dataset $\mathcal{D}$ without using labels. Then, the pretrained GNN is adopted to the downstream classifier. Only a small set of samples $\mathcal{D}_L \subset \mathcal{D}$ with downstream task labels are available for the training of the classifier. During the training of the classifier for downstream tasks, the parameters of pretrained GNN encoder can be fixed or finetuned. Both scenarios of \textit{Fix} and \textit{Finetune} are considered in our experiments. More details about the semi-supervised setting are in Appendix~\ref{sec:addition_setting}. 50 piracy and independent models are built to compute metrics as described in Sec.~\ref{sec:experimental_setup}.
The results of watermarking GraphCL are reported in Tab.~\ref{tab:semi_GraphCL_IP_performance}.  We also report the test accuracy of pretrained GNNs without watermarking as a reference.  From the table, we observe that: 

\begin{itemize}[leftmargin=*]
    \item After fine-tuning the watermarked GNN in downstream tasks, the IP ROC and IP Gap of baselines decrease significantly. It implies the necessity of developing finetuning-resistant watermarking for the pretraining of GNNs.
    \item In both \textit{Fix} and \textit{Finetune} scenarios, PreGIP gives much higher IP ROC and IP Gap than baselines. It indicates the effectiveness of PreGIP in watermarking pretrained GNNs even after finetuning for downstream tasks. 
    \item In both \textit{Fix} and \textit{Finetune} scenarios, {\method} achieves accuracy similar to the pretrained GNN without watermarking. This demonstrates that {\method} can obtain a watermarked GNN encoder with high-quality embeddings to facilitate the downstream tasks.
\end{itemize}

\vskip 0.3em
\noindent \textbf{Watermarking Pretraining for Transfer Learning}. 
We also evaluate the performance of PreGIP in the transfer learning setting of molecular property prediction~\cite{you2020graph}. In the transfer learning setting, we first pretrain the GNN in the ZINC15 dataset that contains 200k unlabeled molecules and then finetune the model in 6 molecular datasets (i.e., Tox21, ToxCast, BBBP, BACE, SIDER, and ClinTox). GraphCL is set as the pretraining strategy. 10 pirated classifiers and independent classifiers are built to compute the IP Gap and IP ROC. All other evaluation settings are the same as Sec.~\ref{sec:experimental_setup}. More details of the transfer learning setting are in Appendix~\ref{appendix:dataset_statistics}. 
The results on the 6 datasets are reported in Tab.~\ref{tab:transfer_GraphCL_IP_performance}. Compared with the baselines, the pretrained GNN encoder watermarked by {\method} can effectively identify piracy models even if the watermarked encoder is updated under the transfer learning setting. Meanwhile, performance on the downstream tasks is comparable with non-watermarked pretrained GNN. This validates the watermarking effectiveness of PreGIP in the transfer learning setting.

\vskip 0.3em
\noindent \textbf{Flexibility of {\method} in Pretraining Strategies}. To show the flexibility of {\method} to different pretraining strategies, we conduct experiments on a representative generative pretraining method, i.e., edge prediction~\cite{hu2020pretraining}. Semi-supervised graph classification is applied as the downstream task. We report IP ROC and accuracy on PROTEINS in Fig.~\ref{fig:flexibility_of_watermarking}. More comparisons on other datasets and baselines can be found in Appendix~\ref{appendix:additional_results_flexibility}.  From the figure, we observe that {\method} consistently achieves better IP ROC and accuracy compared to baselines in both \textit{Fix} and \textit{Finetune} scenarios. It demonstrates the effectiveness of {\method} for the edge prediction pretraining scheme, which implies the flexibility of {\method} to pretraining methods.


\subsection{Resistance to Watermark Removal Methods}
We have demonstrated that {\method} is resistant to the \textit{finetuning} in both semi-supervised and transfer learning settings in Sec.~\ref{sec:results_of_watermarking_pretraining}. The impacts of finetuning epochs to {\method} are also investigated in following. To answer \textbf{Q2}, apart from finetuning, we further investigate the resistance to two widely-used watermark removal methods, i.e., \textit{model pruning}~\cite{zhu2017prune} and \textit{watermark overwriting}~\cite{chen2018performance}.

\begin{figure}[t!]
    \small
    \centering
    \begin{subfigure}{0.48\linewidth}
        \includegraphics[width=\linewidth]{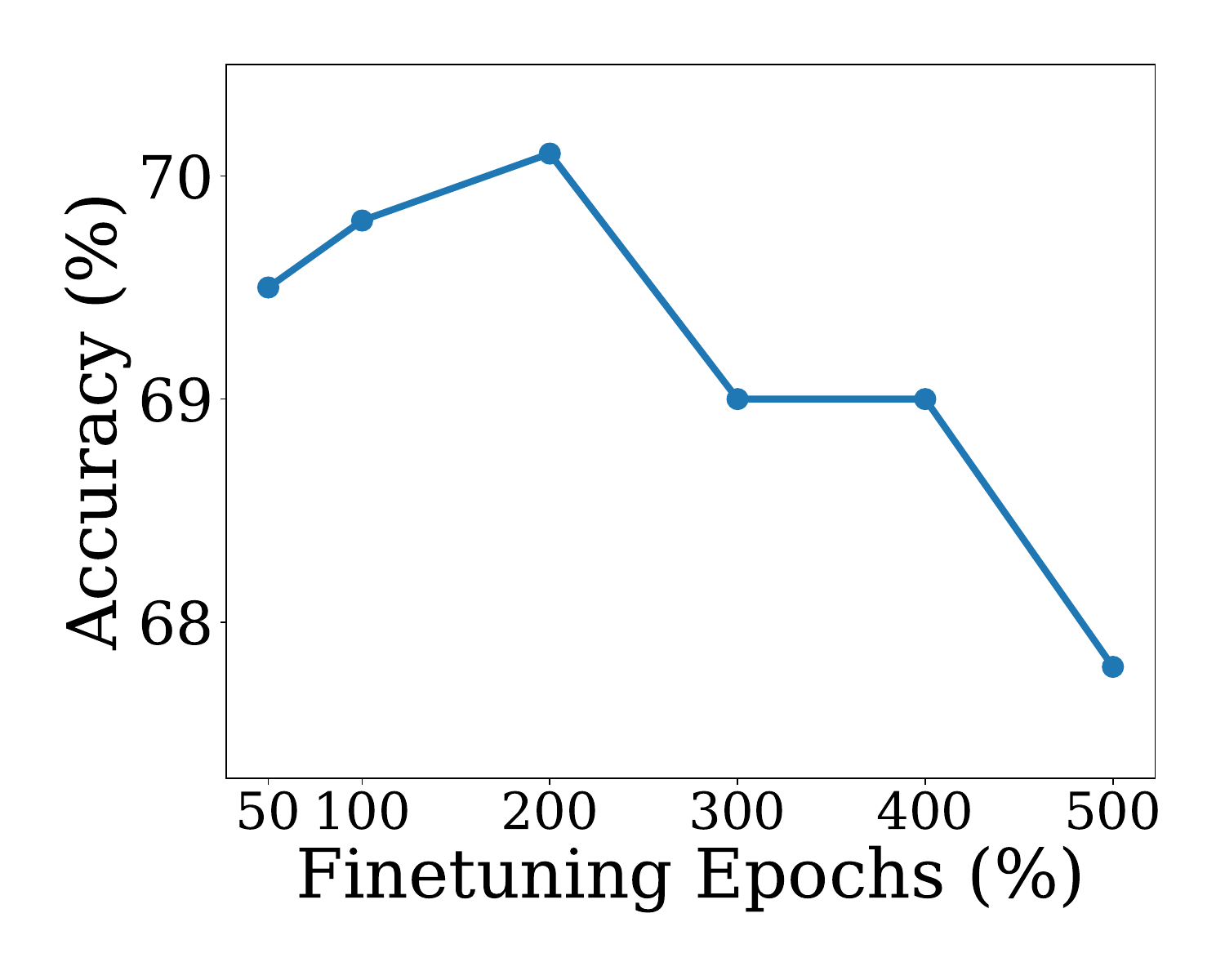}
        \vskip -1em
        \caption{Accuracy}
    \end{subfigure}
    \begin{subfigure}{0.48\linewidth}
        \includegraphics[width=\linewidth]{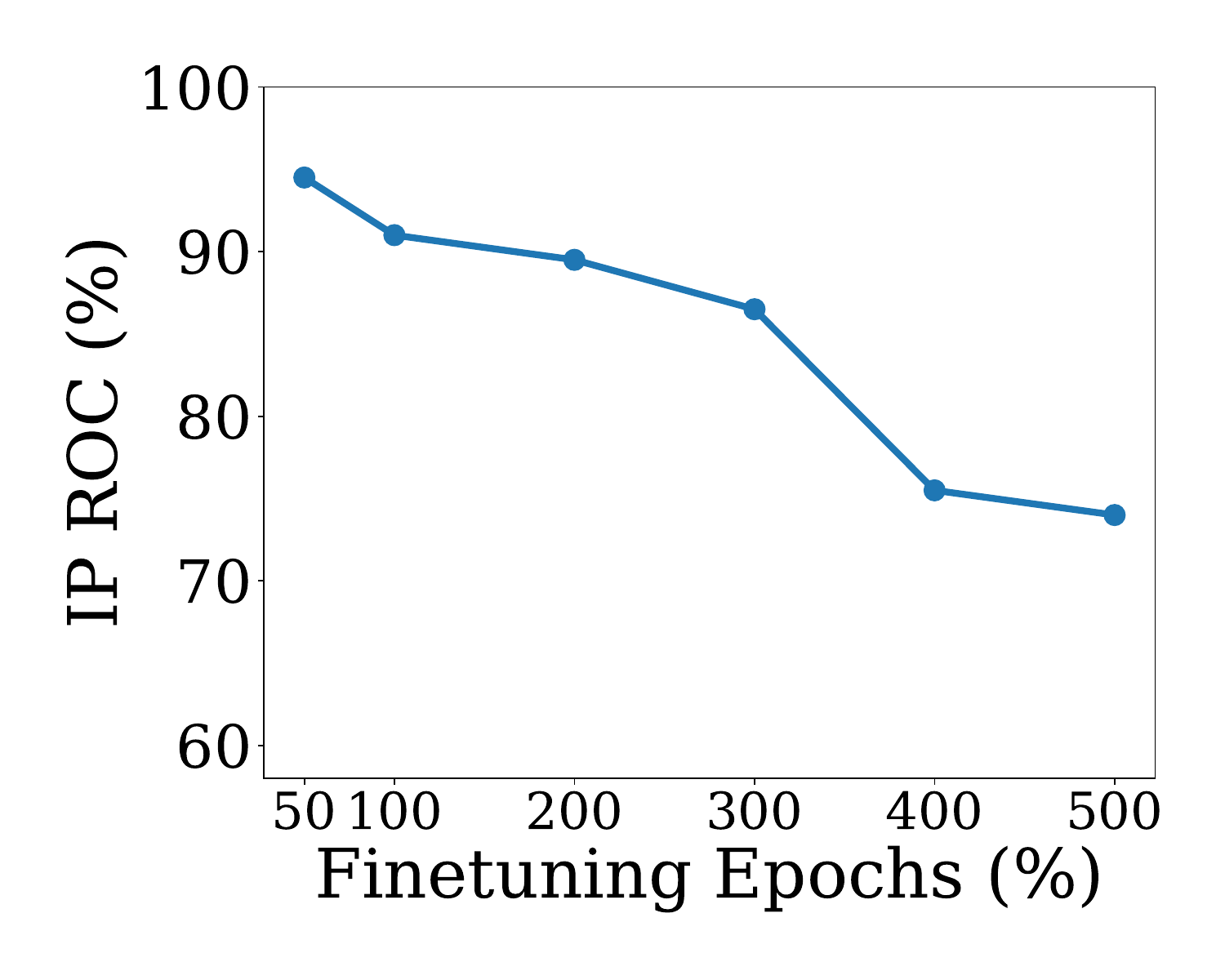}
        \vskip -1em
        \caption{IP ROC }
    \end{subfigure}
    \vskip -1em
    \caption{Impacts of finetuning epochs to {\method}.}
    \vskip -1em
    \label{fig:finetune_epoch}
\end{figure}

\subsubsection{Impacts of Finetuning Epochs}
\label{sec:finetune}
According to Corollary~\ref{theorem:2}, the effectiveness of watermarking is theoretically guaranteed if parameter updates during fine-tuning, i.e., $\delta$ stay below the threshold $\epsilon$.  The $\epsilon$ can be adjusted to tune levels against the fine-tuning. In practice, we can set a value of $\epsilon$ which can balance the utility and watermarking robustness. When long fine-tuning is applied to the watermarked pretraining model, the updates might surpass the threshold $\epsilon$, breaking the watermark. However, the long fine-tuning will also lead to significant overfitting on the labeled set, degrading the performance on downstream tasks. \textit{Hence, it is impractical to break the watermark by long fine-tuning. }

We further conduct experiments to empirically show how different levels of fine-tuning affect the watermarking. Specifically, we vary the epochs of fine-tuning on downstream tasks as \{50, 100, 200, 300, 400, 500\}. The results on PROTEINS are presented in Fig.~\ref{fig:finetune_epoch}. The GNN pretraining strategy is contrastive learning. We have similar observations on other datasets and learning settings. From Fig.~\ref{fig:finetune_epoch}, we can observe that: (i) Watermarks are preserved up to 300 epochs of fine-tuning. (ii) Beyond 300 epochs, watermarks may be compromised, but at the cost of reduced model accuracy due to overfitting. 
The observations are consistent with our analysis that it is impractical to break the watermark by long fine-tuning.

\begin{figure*}[t]
    \small
    \centering
    \begin{subfigure}{0.24\linewidth}
        \includegraphics[width=0.98\linewidth]{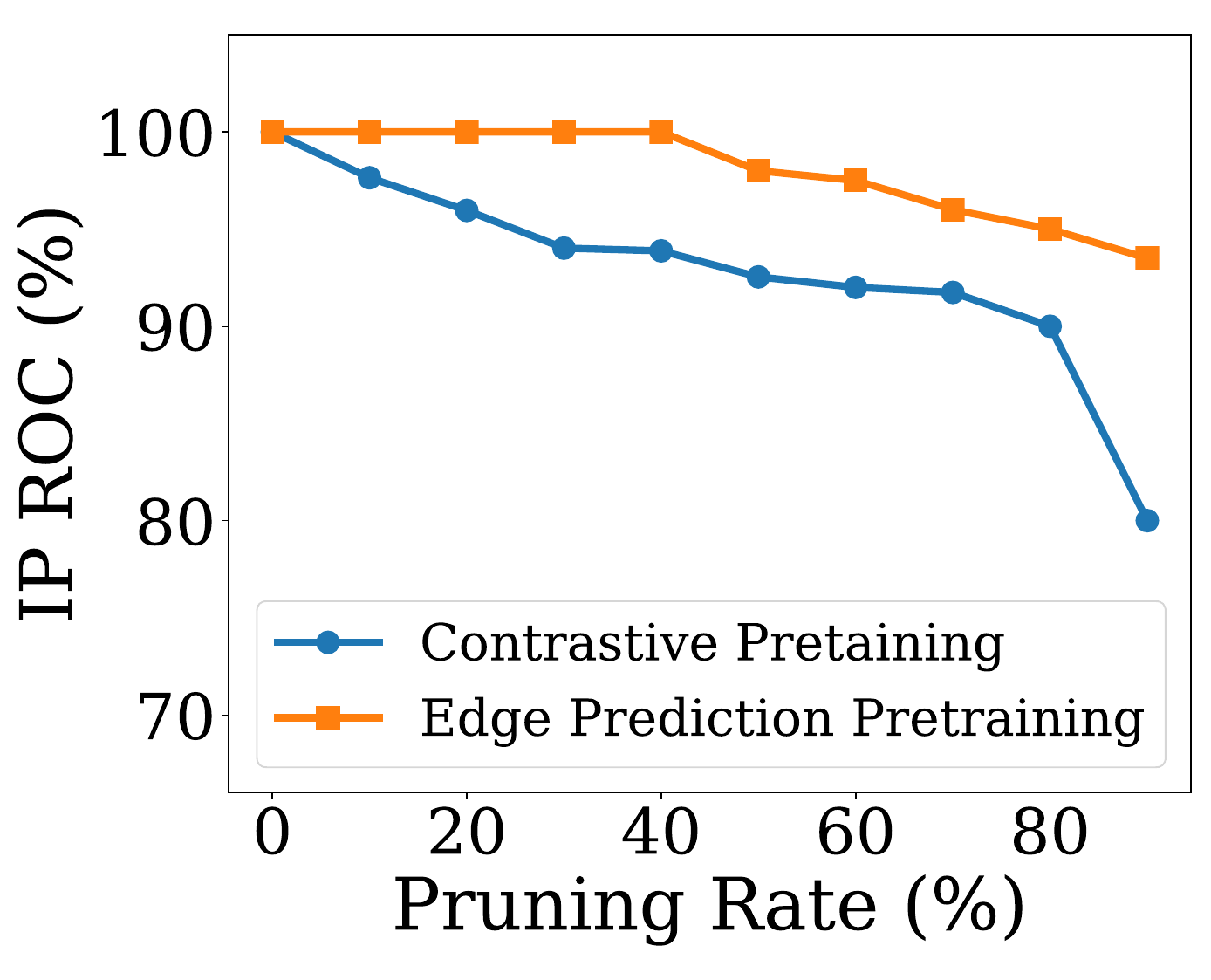}
        \vskip -1em
        \caption{IP ROC on PROTEINS}
    \end{subfigure}
    \begin{subfigure}{0.24\linewidth}
        \includegraphics[width=0.98\linewidth]{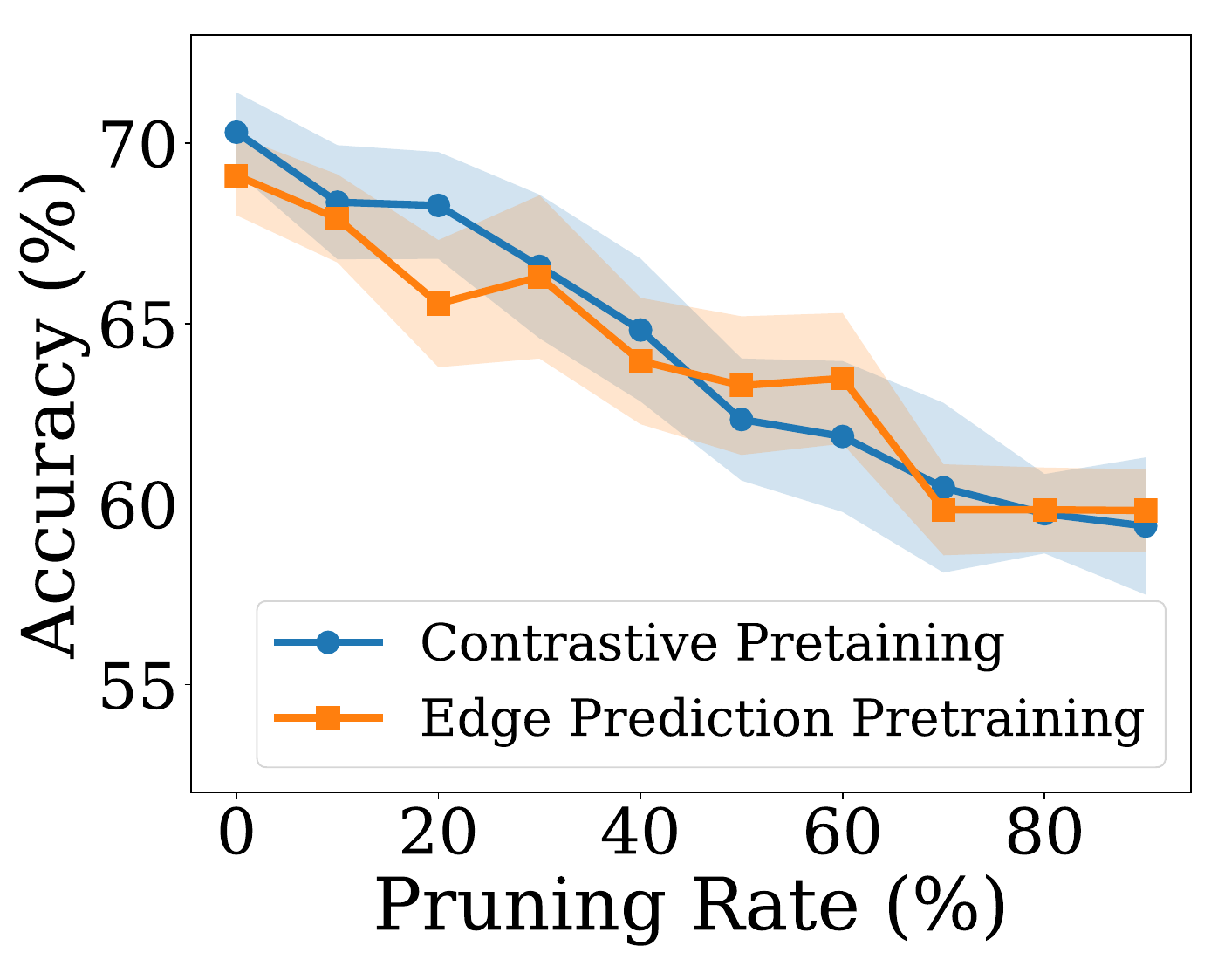}
        \vskip -1em
        \caption{Accuracy on PROTEINS}
    \end{subfigure}
    \begin{subfigure}{0.24\linewidth}
        \includegraphics[width=0.98\linewidth]{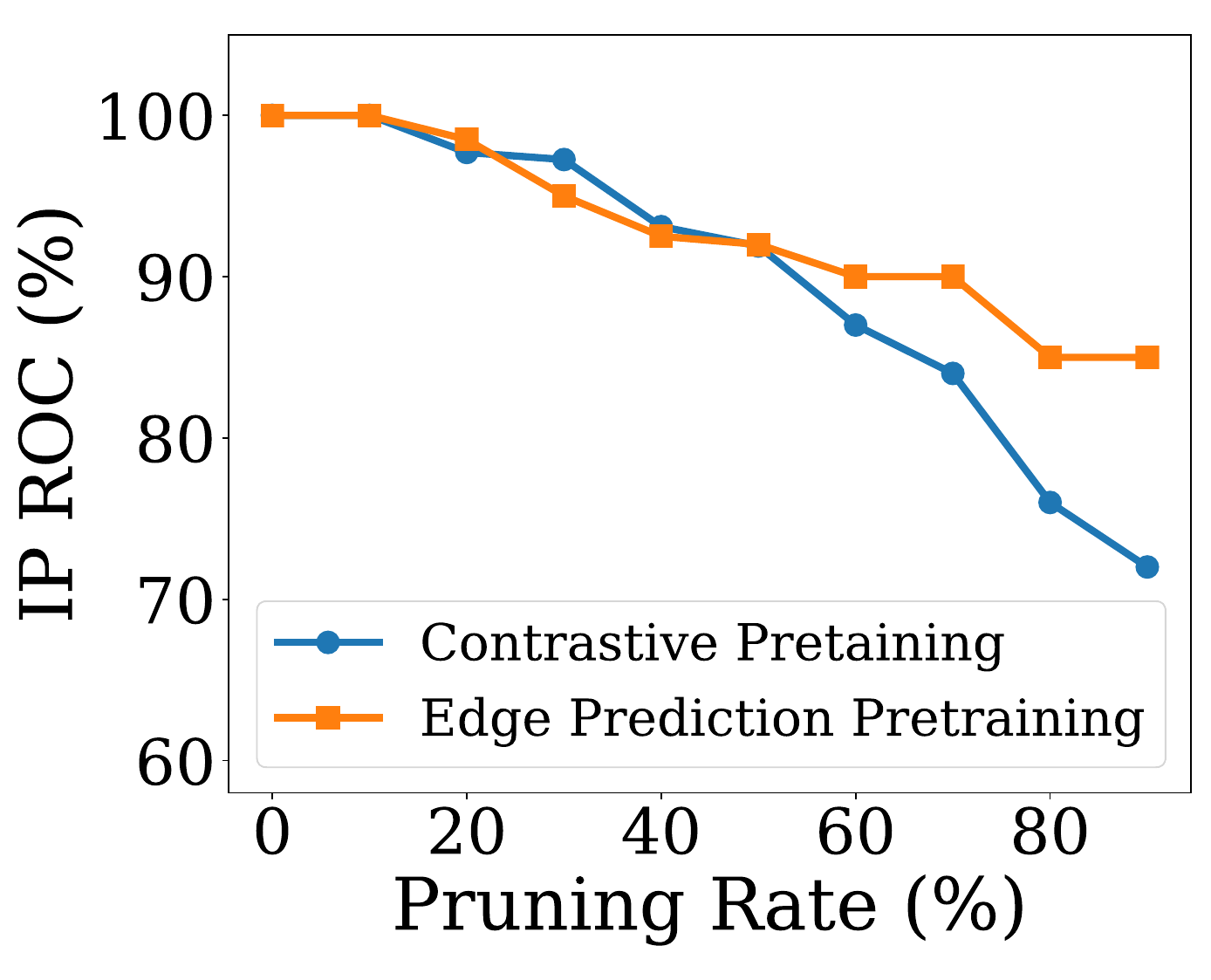}
        \vskip -1em
        \caption{IP ROC on NCI1}
    \end{subfigure}
    \begin{subfigure}{0.24\linewidth}
        \includegraphics[width=0.98\linewidth]{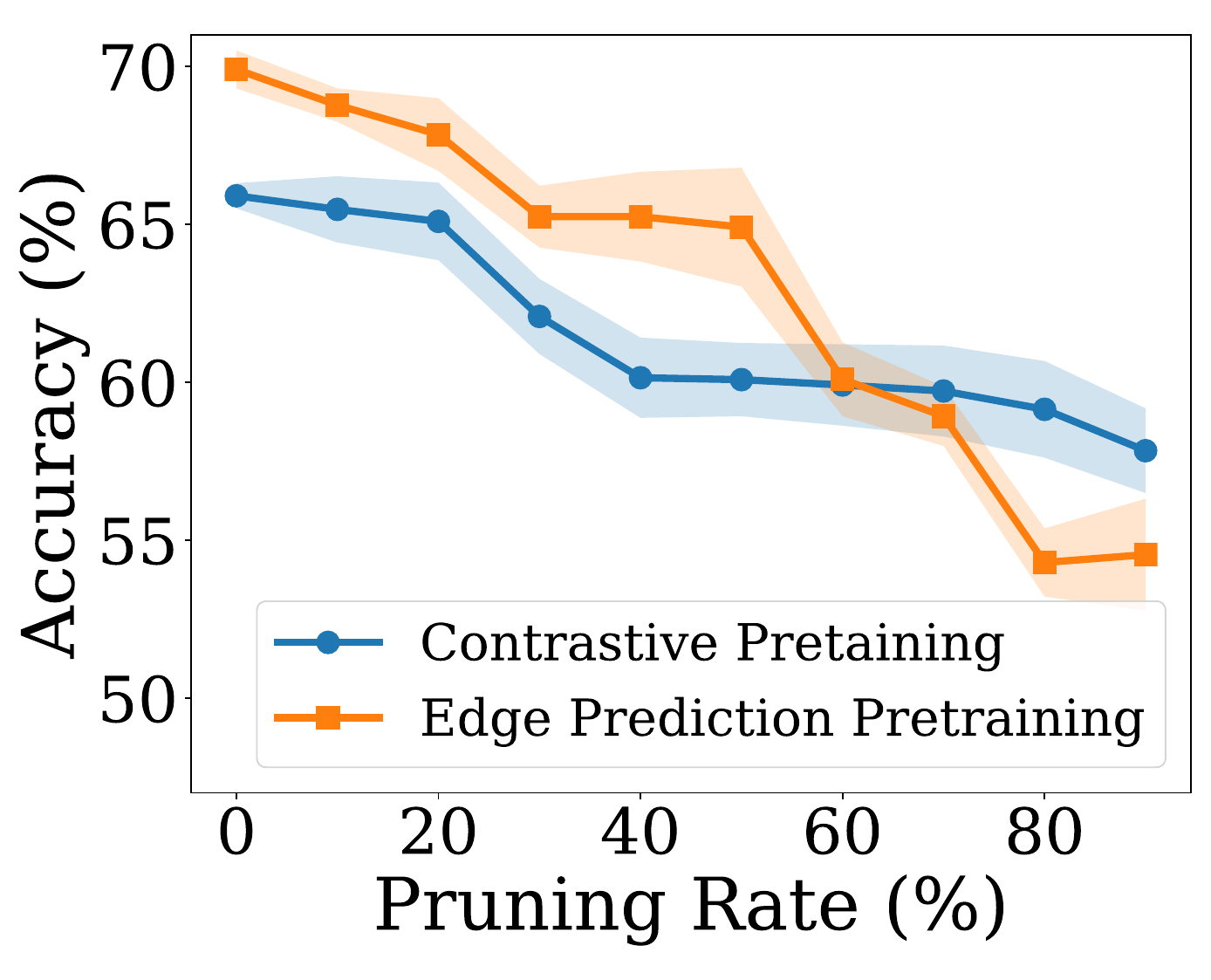}
        \vskip -1em
        \caption{Accuracy on NCI1}
    \end{subfigure}
    \vskip -0.5em
    \caption{Impacts of pruning rates to watermarking performance and downstream task accuracy.}
    \label{fig:pruning}
\end{figure*}

\begin{figure}[t!]
    \small
    \centering
    \begin{subfigure}{0.48\linewidth}
        \includegraphics[width=\linewidth]{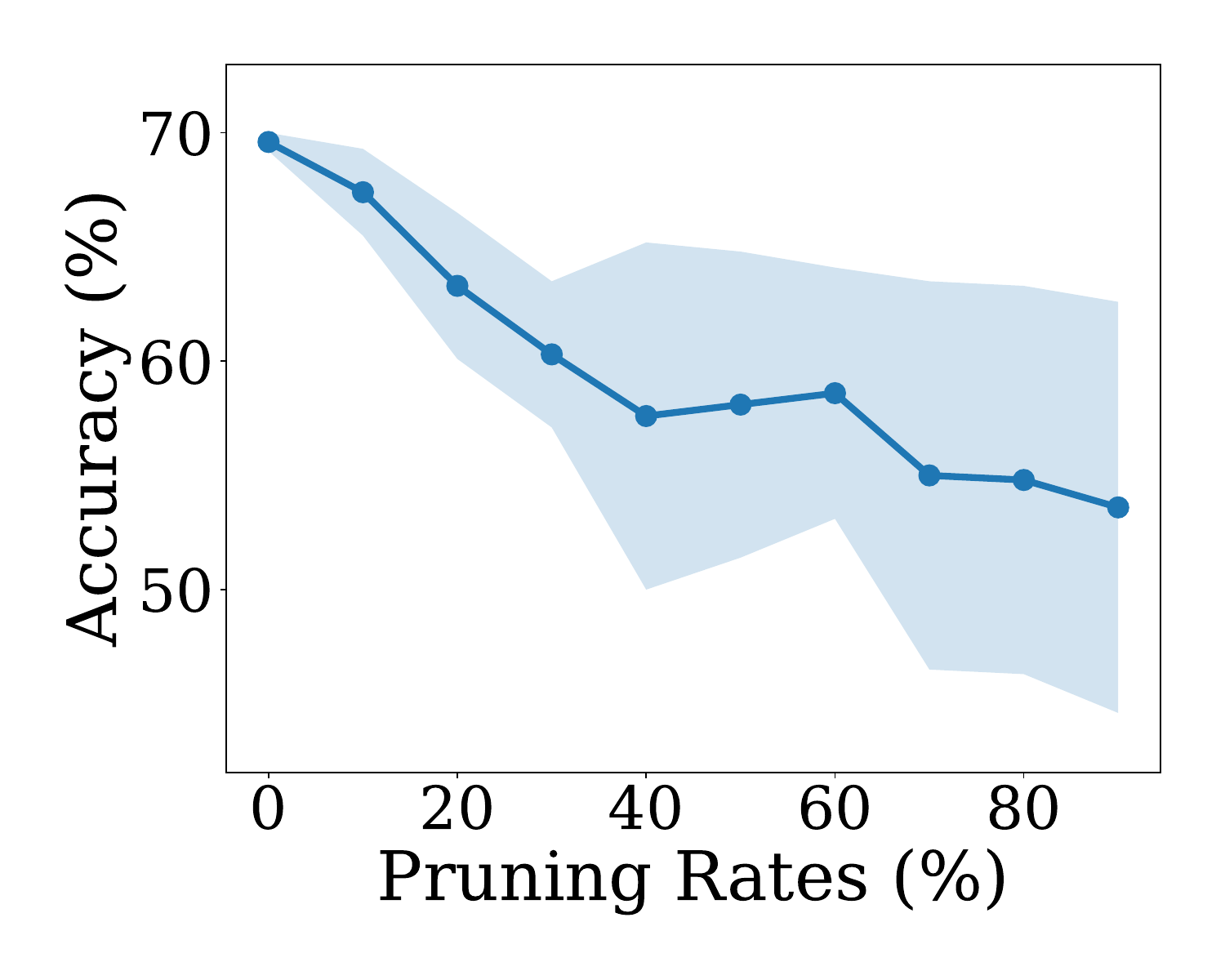}
        \vskip -0.5em
        \caption{Accuracy}
    \end{subfigure}
    \begin{subfigure}{0.48\linewidth}
        \includegraphics[width=\linewidth]{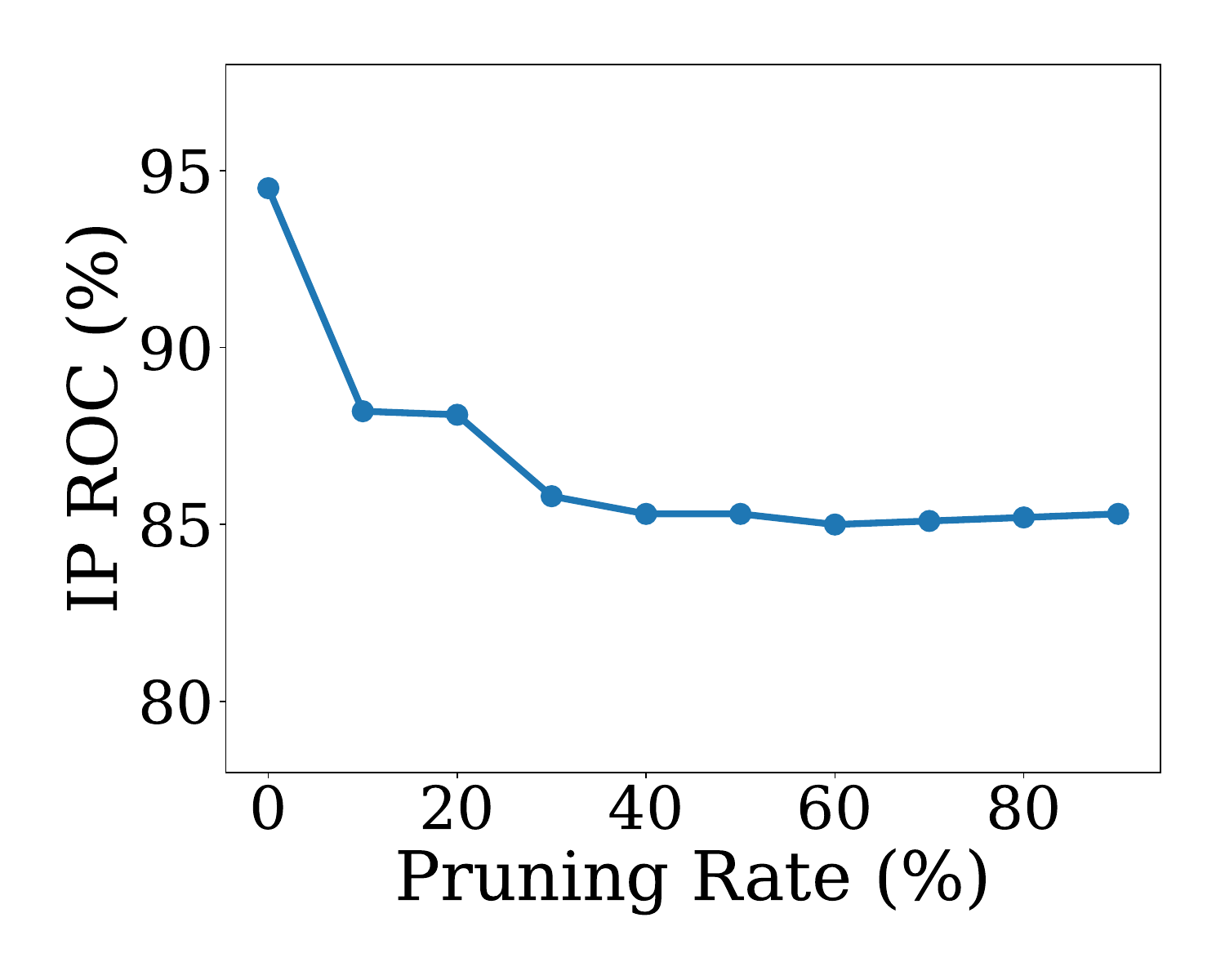}
        \vskip -0.5em
        \caption{IP ROC}
    \end{subfigure}
    \vskip -0.5em
    \caption{Results of PreGIP against combination attack in which both finetuning and pruning are applied.}
    \label{fig:combination}
\end{figure}

\subsubsection{Model Pruning} 
In model pruning, neurons will be removed for compression, which can invalidate watermarks~\cite{uchida2017embedding}. Therefore, following~\cite{wu2022watermarking,uchida2017embedding}, we evaluate the effectiveness of {\method} under model pruning~\cite{zhu2017prune}. Specifically, we vary the rates of pruning the watermarked GNN model parameters from 10\% to 90\% with a step of 10\%.  Experiments are conducted on two pretraining strategies, i.e., watermarking contrastive pretraining and edge prediction pretraining. The results are shown in Fig.~\ref{fig:pruning}. We can observe that even when the pruning rate reaches 50\% which leads to a poor classifier, the IP ROC remains higher than 90\% in both datasets.  This demonstrates that {\method} is resistant to the model pruning. 

We further testify a combination attack method which combines both finetuning and model pruning to break the watermark. Specifically, the watermark model will firstly finetune for 100 epochs. Then, the model pruning would be adapted. In the experiments, we very the pruning ratio from 10\% to 90\% with a step of 10\%. The results on PROTEINS are presented in Fig.~\ref{fig:combination}. The GNN pretraining strategy is contrastive learning. We have similar observations on other datasets and learning settings. From Fig.~\ref{fig:combination}, we can find that the IP ROC remains higher than 85\% even with both finetuning and pruning strategies. This demonstrates that PreGIP is resistant to the combination attack method.

\subsubsection{Watermark Overwriting}
The overwriting aims to embed new watermarks to render original watermarks undetectable~\cite{chen2018performance}. In our experiments, the adversary simultaneously injects their watermark pairs and finetunes the watermarked GNN. All other experimental settings are kept the same as Sec.~\ref{sec:experimental_setup}. 
The results are presented in Tab.~\ref{tab:overwriting}. From this table, we can find the IP ROC maintains higher than 90\% after the overwriting attack. It indicates the effectiveness of our PreGIP against watermark overwriting.

\begin{table}[h]
    \centering
    \small
    \caption{Results of PreGIP under Watermark Overwriting.}
    \vskip -1em
    \begin{tabularx}{0.95\linewidth}{XCCC}
    \toprule
     Dataset  & Accuracy (\%) &IP GAP (\%) & IP ROC (\%) \\
    \midrule
    PROTEINS & 69.8 $\pm 0.8$ & 42.5 $\pm 29.6$ & 93.0 \\
    NCI1 & 68.5 $\pm 1.1$ & 25.0 $\pm 0.9$ & 99.0 \\
    \bottomrule
    \end{tabularx}
    \label{tab:overwriting}
    \vskip -1em
\end{table}

\subsection{Impact of the Number of Watermark Graphs}
\label{sec:impact_of_the_number_watermark_graph_pairs}
\begin{figure}[h]
    \small
    \centering
    \begin{subfigure}{0.49\linewidth}
        \includegraphics[width=0.92\linewidth]{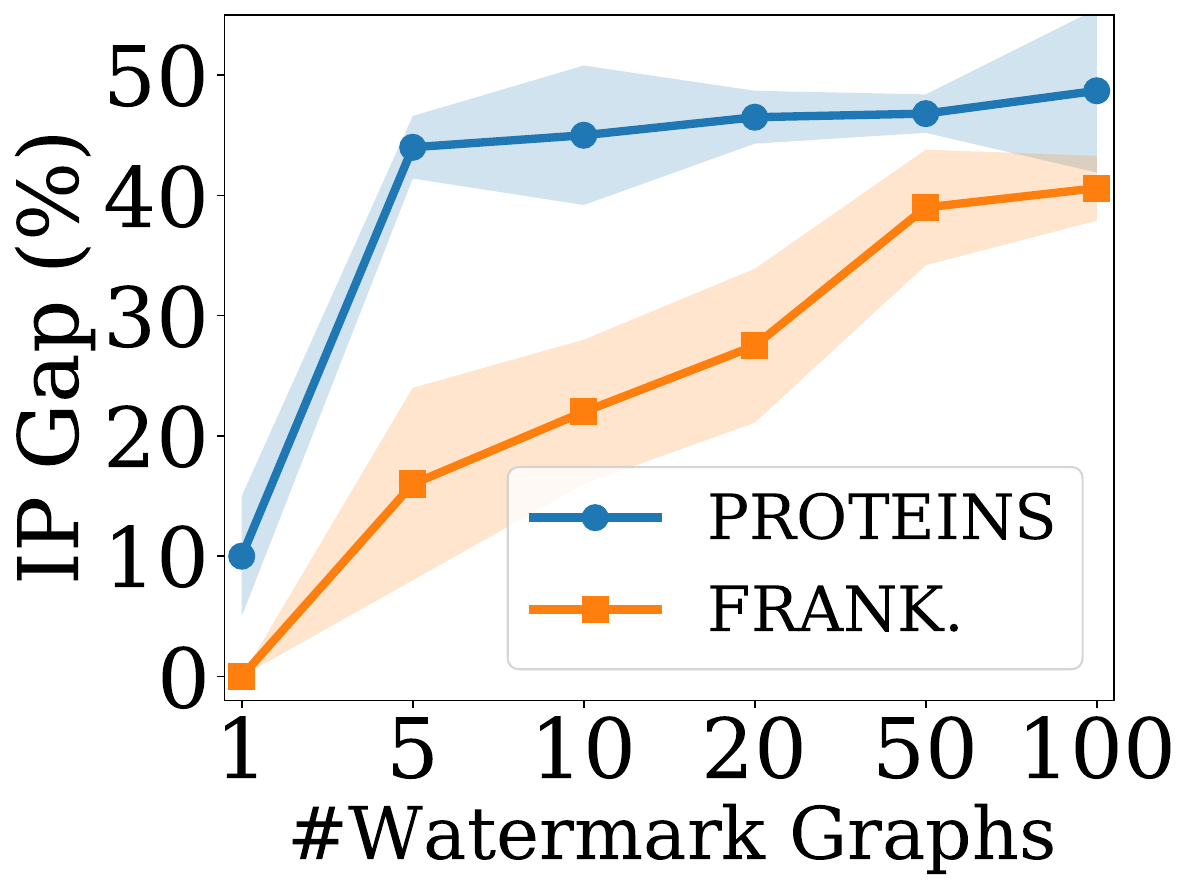}
        \vskip -0.5em
        \caption{IP Gap}
    \end{subfigure}
    \begin{subfigure}{0.49\linewidth}
        \includegraphics[width=0.92\linewidth]{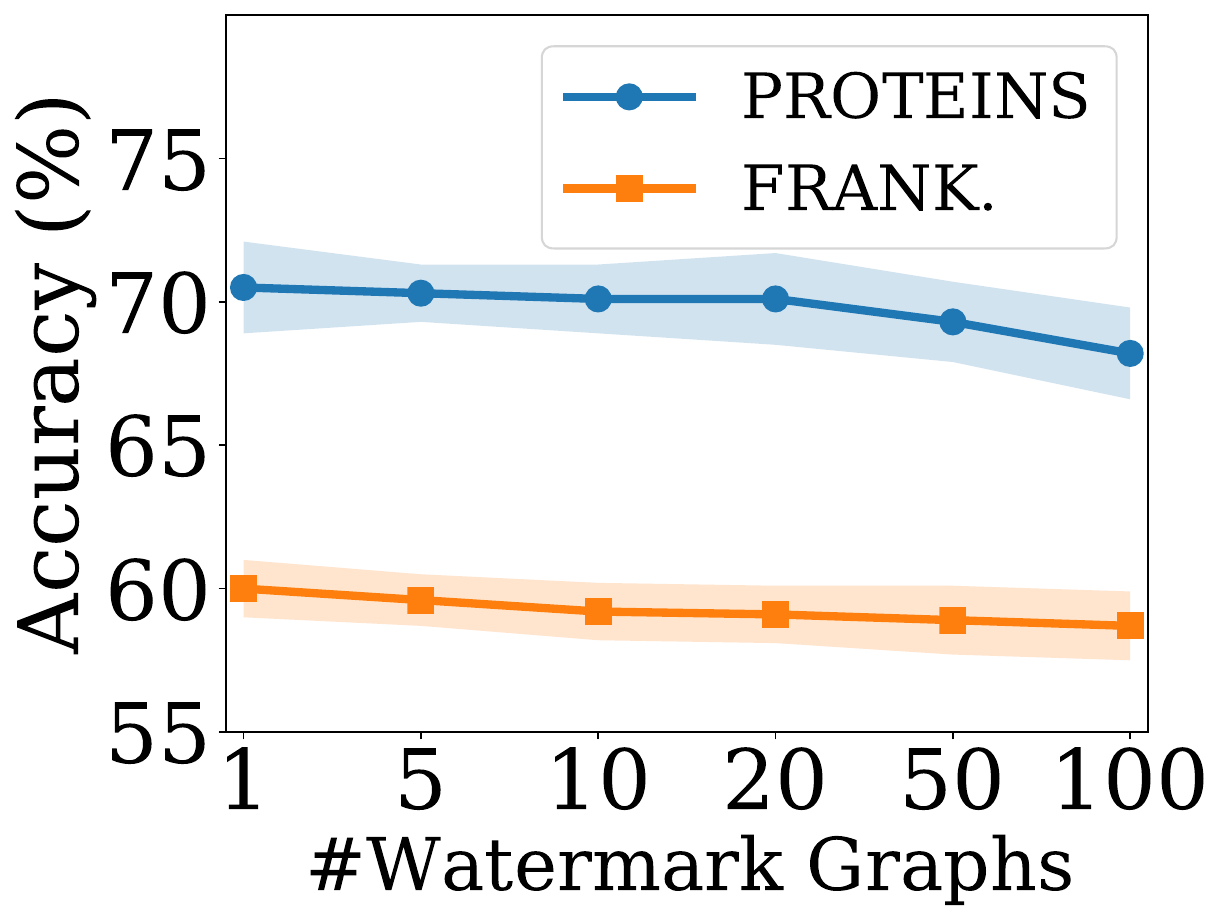}
        \vskip -0.5em
        \caption{Accuracy}
    \end{subfigure}
    \vskip -1em
    \caption{Impacts of the number of watermark graph pairs.}
    \vskip -1em
    \label{fig:impact_of_the_budget}
\end{figure}

We conduct experiments to explore the watermarking performance of PreGIP given different budgets in the number of watermark graph pairs. Specifically, we vary the number of watermarking graphs as $\{1,5,10,20,50,100\}$. We set GraphCL as the pretraining scheme and the semi-supervised graph classification as the downstream task. The other settings are described in Sec.~\ref{sec:experimental_setup}. Fig.~\ref{fig:impact_of_the_budget} reports the IP Gap and accuracy on PROTEINS and FRANKENSTEIN. we can observe that as the number of watermarking graphs increases, the IP Gap consistently rises, while accuracy slightly decreases with more watermark graph pairs in training. This suggests more watermark graph pairs can strengthen the watermarking, but too many may negatively affect the embedding space of the pretrained GNN. Hence, an appropriate number of watermark graph pairs is necessary to balance the IP protection and model utility.

\subsection{Ablation Studies}
\begin{figure}[h]
    \small
    \centering
    \vskip -1em
    \begin{subfigure}{0.49\linewidth}
        \includegraphics[width=0.95\linewidth]{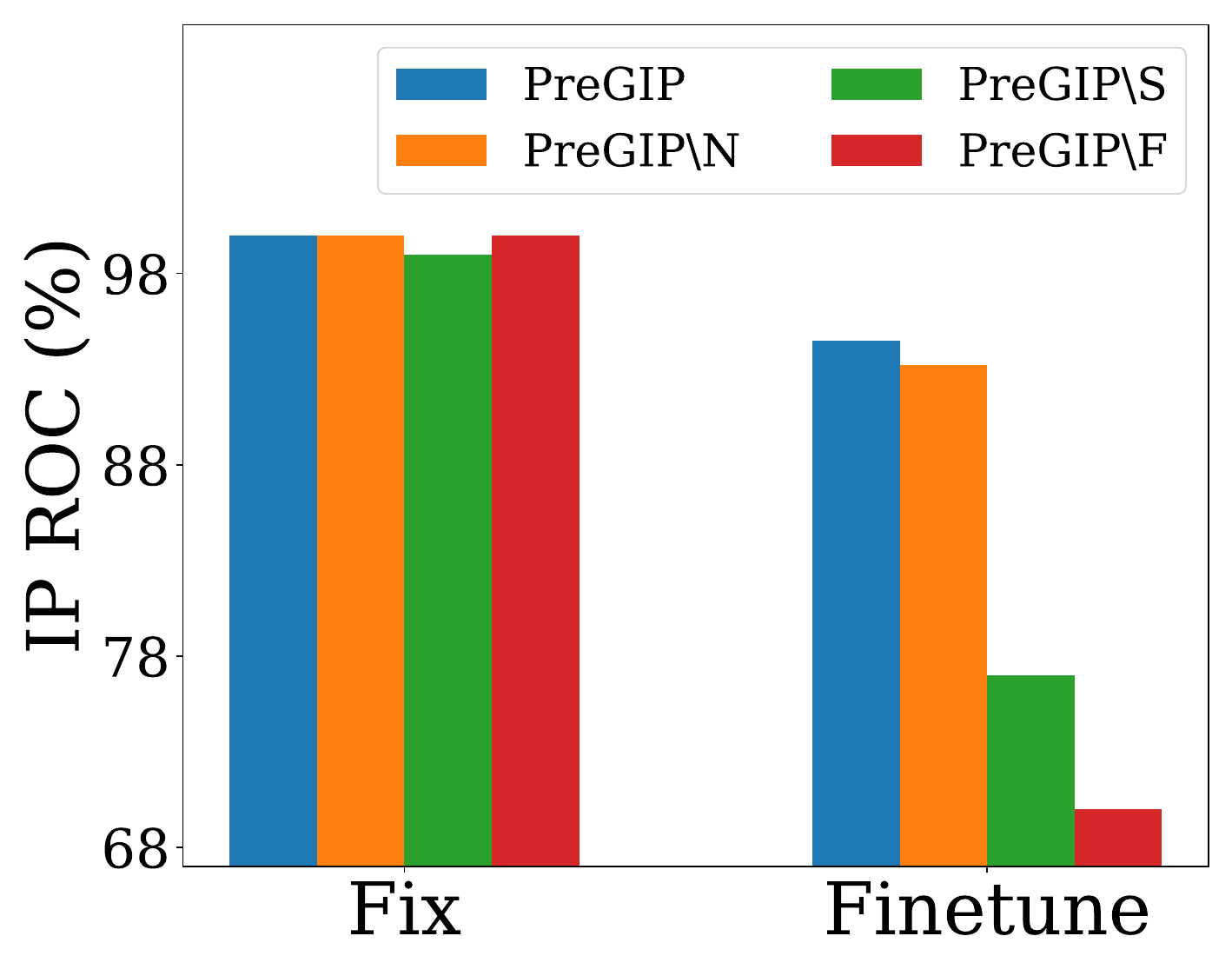}
        \vskip -0.5em
        \caption{IP ROC}
    \end{subfigure}
    \begin{subfigure}{0.49\linewidth}
        \includegraphics[width=0.95\linewidth]{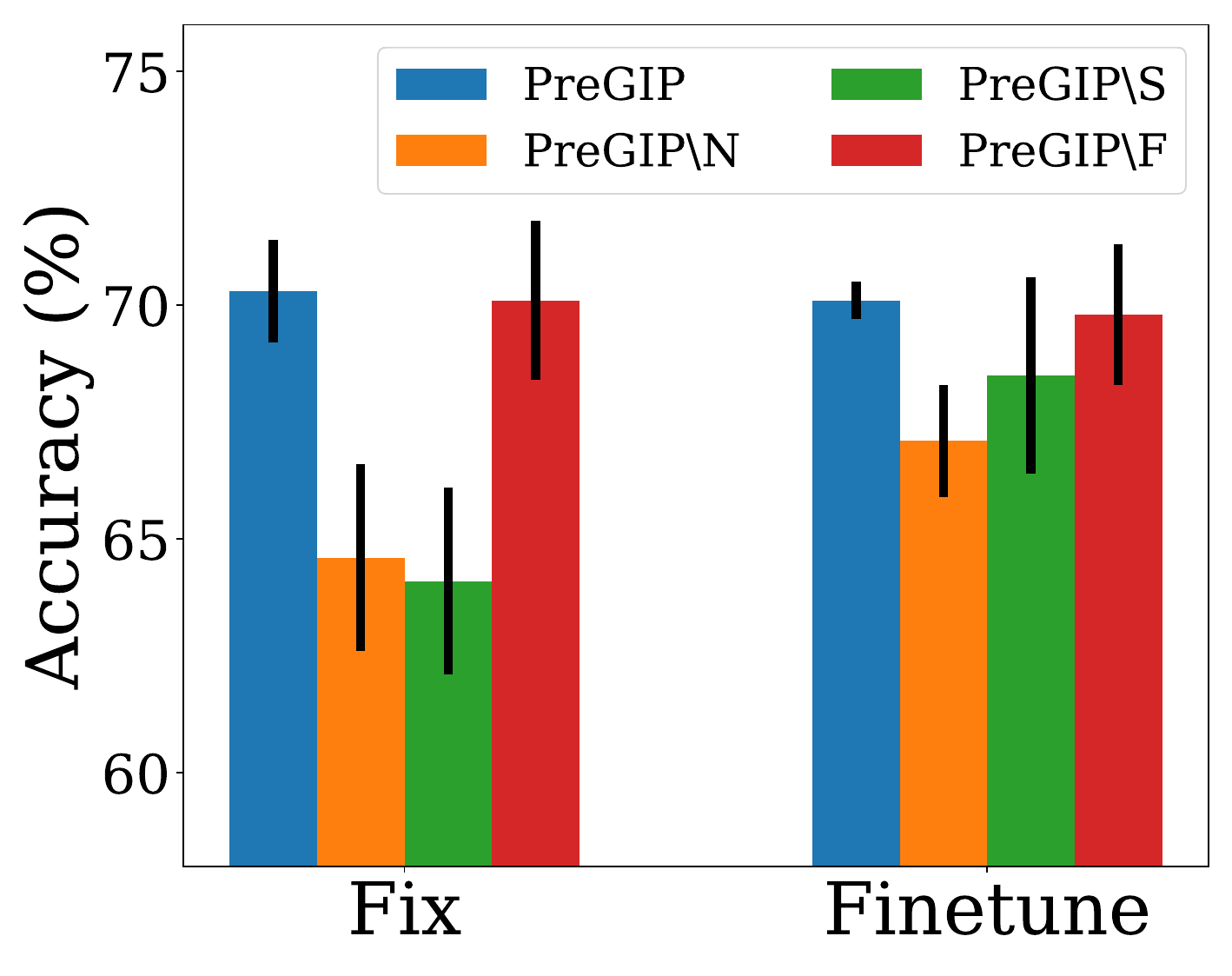}
        \vskip -0.5em
        \caption{Accuracy}
    \end{subfigure}
    \vskip -1.5em
    \caption{Ablation study results on PROTEINS.}
    \vskip -1.5em
    \label{fig:ablation_studies_NCI1}
\end{figure}

To answer \textbf{Q3}, we conduct ablation studies to understand the effects of the proposed watermarking loss and the finetuning-resistant watermarking. 
To show quality of embeddings can be preserved with our proposed watermarking loss, we train a variant {\method}$\backslash$N that only ensures high similarity for paired watermark graphs without the latter term in Eq(\ref{eq:watermark}).  {\method} deploys synthetic graphs as watermark graphs to reduce negative effects to real-world graph embeddings. To prove its effectiveness, we randomly sample real graphs from pretraining set as watermark graphs and train a variant named {\method}$\backslash$S. Moreover, a variant named {\method}$\backslash$F is trained without the finetuning-resist watermarking mechanism to show the contribution of this mechanism. The IP protection results and performance in downstream tasks are given in Fig.~\ref{fig:ablation_studies_NCI1}. Specifically, we observe that:
\begin{itemize}[leftmargin=*]
    \item Both {\method}$\backslash$N and {\method}$\backslash$S perform significantly worse than {\method} in terms of downstream classification accuracy. This demonstrates the proposed watermarking loss in Eq.(\ref{eq:watermark}) and the deployment of synthetic watermark graphs can effectively preserve the quality of watermarked GNN encoder's embeddings.
    \item {\method} outperforms {\method}$\backslash$F by a large margin in IP ROC when the watermarked GNN encoder is finetuned for downstream tasks. This proves the effectiveness of our proposed finetuning-resist watermark injection. 
\end{itemize}


\subsection{Hyperparameter Analysis} \label{sec:hyper}

\begin{figure}[h]
    \small
    \centering
    \begin{subfigure}{0.49\linewidth}
        \includegraphics[width=\linewidth]{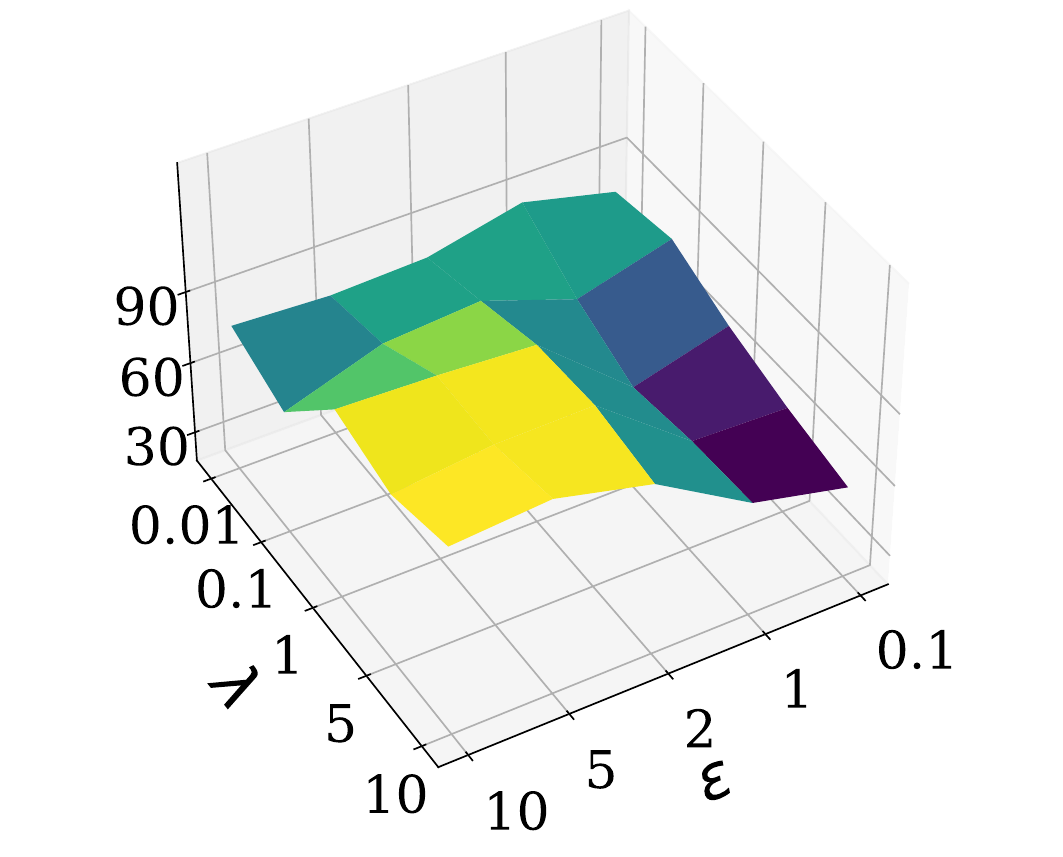}
        \vskip -0.5em
        \caption{IP ROC (\%)}
    \end{subfigure}
    \begin{subfigure}{0.49\linewidth}
        \includegraphics[width=\linewidth]{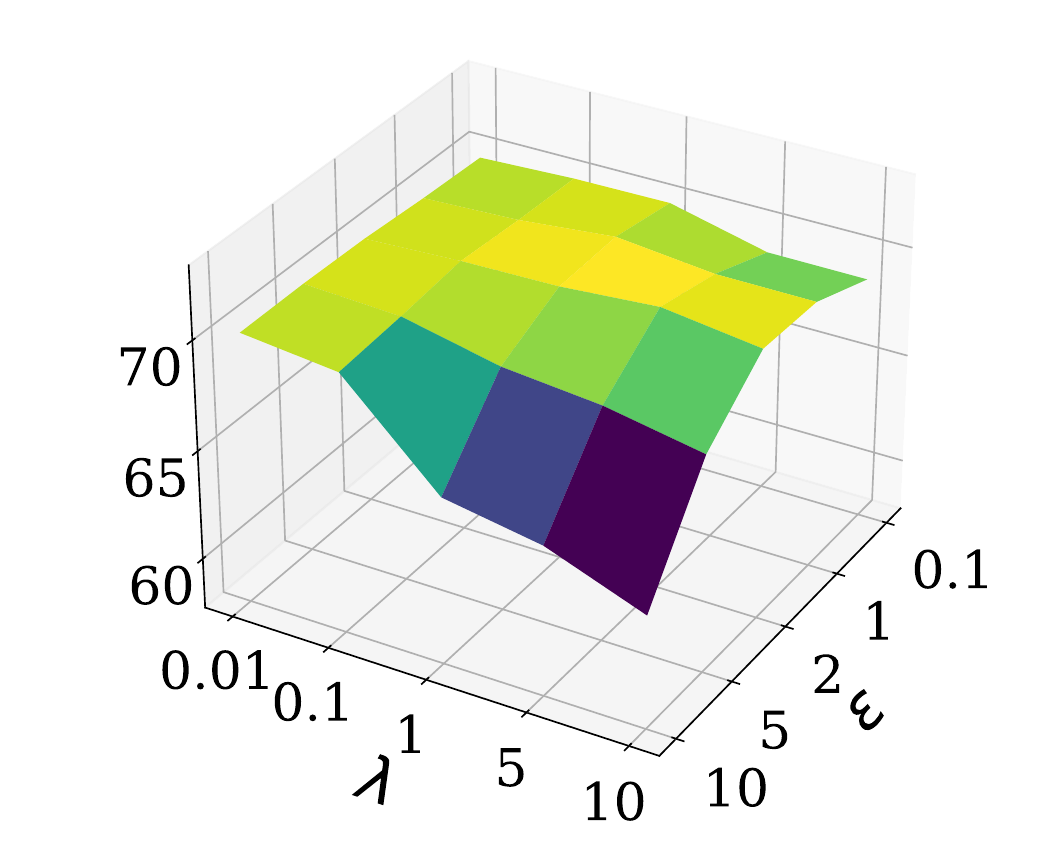}
        \vskip -0.5em
        \caption{Accuracy (\%)}
    \end{subfigure}
    \vskip -1.2em
    \caption{Hyperparameter analysis on PROTEINS.}
    \vskip -1.5em
    \label{fig:hyper}
\end{figure}

We further investigate how hyperparameters $\lambda$ and $\epsilon$ affect the performance of PreGIP, where $\lambda$ controls the contributions of the proposed watermarking loss on the pretrained models, and $\epsilon$ controls the magnitude of changes to the model parameters $\theta$ in finetuning-resist watermark injection. More specifically, we vary $\lambda$ and $\epsilon$ as $\{0.01,0.1,1,5,10\}$ and $\{0.1,1,2,5,10\}$, respectively. We report the IP ROC and accuracy on PROTEINS datasets. The experimental setting is semi-supervised under the finetuning scenario. 
The results are shown in Fig.~\ref{fig:hyper}. From this figure, we find that IP ROC will increase with the increase of $\lambda$ and $\epsilon$. As for the accuracy of the watermarked GNN encoder in downstream tasks, it will first stay consistent and then decrease if the $\lambda$ and $\epsilon$ are overly high.  
According to the hyperparameter analysis on PROTEINS, we set $\epsilon$ and $\lambda$ as 2 and 1. Then, the same hyperparameter setting is applied to all other datasets and experimental settings without further tuning. 

\section{Related Works}

\textbf{Graph Model Pretraining}. The pretraining on graph neural networks aims to learn useful knowledge from large-scale data to facilitate various downstream tasks~\cite{you2020graph,sun2023all,dai2021towards,dai2025towards}. 
During the pretraining phase, a pretext task such as graph construction will be deployed to train the GNN encoder for a high-quality embedding space. 
Based on the types of pretext tasks, graph pretraining methods can be categorized into contrastive pretraining and generative pretraining. Graph contrastive learning~\cite{you2020graph,qiu2020gcc,lin2023certifiably} aims to maximize the mutual information between different augmentation views of graphs. For instance, GraphCL~\cite{you2020graph} testifies to the pretraining performance with different combinations of augmentation views. Recently, All in One~\cite{sun2023all} designs a multi-task prompting framework to unify node-level, edge-level, and graph-level tasks in a pretrained contrastive learning model. The generative pretraining aims to train the graph model by generative tasks such as graph reconstruction and attribute prediction. For example, edge prediction and masked node prediction are investigated in~\cite{hu2020pretraining}. GraphMAE~\cite{hou2022graphmae} deploys the feature reconstruction with a re-mask decoding strategy.

\noindent \textbf{Model Watermarking}. Apart from robust GNNs~\cite{dai2021nrgnn,dai2022towards,dai2023unified,dai2023unnoticeable,hou2024adversarial}, IP protection of GNNs is also an important aspect of safety~\cite{dai2024comprehensive}. 
Various watermarking methods~\cite{uchida2017embedding,zhang2018protecting,adi2018turning,rouhani2018deepsigns,sun2023deep} are designed for deep IP protection. 
\cite{uchida2017embedding} firstly regularizes model parameters to inject the watermark. 
To achieve black-box verification without using model parameters, backdoor-based watermark methods~\cite{zhang2018protecting,adi2018turning} are proposed to predict trigger samples as a predefined class for watermarking. 
Recently, SSLGuard~\cite{cong2022sslguard} introduced a white-box watermark method for pretrained encoders. Black-box IP watermarking methods~\cite{wu2022watermarking,lv2022ssl,peng2023you,shen2021backdoor} for self-supervised learning on images and language models are further investigated. For instance, DuFFin~\cite{yan2025duffin} deploys a dual-level non-invasive watermarking (fingerprinting) framework for large language models. 
For the graph-structured data, some initial efforts employ watermarking to protect the IP of supervised GNN classifier~\cite{xu2023watermarking}. Following backdoor-based methods on images~\cite{adi2018turning}, ~\cite{xu2023watermarking} enforces the GNN classifier to predict trigger graphs as a certain class to achieve watermarking. Our {\method} is inherently different from the aforementioned methods: (i) we focus on watermarking the pretraining of the GNN  without the downstream classification task information; (ii) we propose a novel finetuning-resistant watermark injection framework, which can identify unauthorized adoption of pretraining GNN encoder even after finetuning on downstream tasks.  (iii) We conduct theoretical analysis to verify the feasibility of {\method}.

\section{Conclusion and Future Work} 
In this paper, we study a novel problem of watermarking the pretraining of GNNs for deep IP protection. Specifically, we propose a task-free model verification framework that can watermark the GNN pretraining without the downstream task information. A watermarking loss is applied to embed the watermarks without degrading the quality of representations obtained by the GNN encoder. Moreover, a finetuning-resistant watermarking approach is implemented to guarantee the identification of any suspect model derived from finetuning the protected GNN encoder. Extensive experiments on different datasets and settings demonstrate that  {\method} can effectively watermark the pretraining of GNNs. In the future, it would be promising to investigate potential watermark removal strategies to bypass the identification of pirated models.

\section{Acknowledgments}
This material is based upon work supported by, or in part by the Army Research Office (ARO) under grant number W911NF-21-10198, the Department of Homeland Security (DHS) under grant number 17STCIN00001-05-00, and Cisco Faculty Research Award. 



\bibliographystyle{ACM-Reference-Format}
\bibliography{ref}

\appendix
\section{Proofs}
\subsection{Proof of Theorem~\ref{theorem:1}}
\label{app:proof1}

The Lipschitz constant of MLP is used to prove Theorem~\ref{theorem:1}. Therefore, we first introduce the definition of the Lipschtiz constant followed by a proposition of computing the upper bound of the Lipschitz constant for MLP. Finally, the proof of Theorem~\ref{theorem:1} can be given. 
\begin{definition}[Lipschitz Constant \cite{virmaux2018lipschitz}] A function $f:\mathbb{R}^n \rightarrow \mathbb{R}^m$ is called Lipschtiz continuous if there exists a real constant $L$ such that:
\begin{equation}
    \forall x_1,x_2 \in \mathbb{R}^n, \|f(x_1) - f(x_2)\|_2 \leq L \| x_1 - x_2\|_2.
\end{equation}
The smallest $L$ for which the above inequality is true is called the Lipschitz constant of $f$.    
\end{definition}

\begin{proposition} \label{proposition} \cite{virmaux2018lipschitz}
    For a $K$-layer MLP with 1-Lipschitz activation functions (e.g., ReLU, sigmoid, tanh, etc.), the upper bound of its Lipschitz constant is:
    \begin{equation}
        L = \prod_{i=1}^K \|\mathbf{W}_i\|_2 
    \end{equation}
    where $\mathbf{W}_i$ denotes the parameters of $i$-th layer in the MLP.
\end{proposition}

With the above definition and proposition, we give the proof to Theorem~\ref{theorem:1} below.
\begin{proof}
    Let $\mathbf{y}_a$ and $\mathbf{y}_b$ represent the unnormalized prediction vector from the $f_C$ on $\mathcal{G}_w^a$ and $\mathcal{G}_w^b$. The margin $s_a$ between the logit scores of the predicted class $c_1$ and the second confident class $c_2$ can be computed by
    \begin{equation}
        s_a = \mathbf{y}_a(c_1) - \mathbf{y}_a(c_2)
    \end{equation}
    With the Proposition~\ref{proposition} and the condition specified in Theorem~\ref{theorem:1} i.e.,  $\|f_E(\mathcal{G}_w^a) - f_E(\mathcal{G}_w^b) \|_2 < \frac{1}{2} \cdot \frac{s_a}{\prod_{i=1}^K \|\mathbf{W}_i\|_2}$, we can obtain
    \begin{equation}
        \| \mathbf{y}_a - \mathbf{y}_b\|_2 \le L \|f_E(\mathcal{G}_w^a) - f_E(\mathcal{G}_w^b) \|_2 < \prod_{i=1}^K \|\mathbf{W}_i\|^2 \cdot \frac{1}{2} \cdot \frac{s_a}{\prod_{i=1}^K \|\mathbf{W}_i\|^2} = \frac{1}{2} s_a
    \end{equation}
    With $\|\mathbf{y}_b-\mathbf{y}_a\|_2 < \frac{1}{2} \cdot s_a$, we know that the largest difference between $\mathbf{y}_a$ and $\mathbf{y}_b$ in each class would be less than $\frac{1}{2} s_a$, i.e.,
    \begin{equation}
        |\mathbf{y}_b(i) - \mathbf{y}_a(i)| < \frac{1}{2} s_a
    \end{equation}
    where $\mathbf{y}_a(i)$ denotes the $i$-th logit of $\mathbf{y}_a$. Hence, we have $\mathbf{y}_a(i) - \frac{1}{2} s_a < \mathbf{y}_b(i) < \mathbf{y}_a(i) + \frac{1}{2} s_a$. We then can get
    \begin{equation}
        \begin{aligned}
            \mathbf{y}_b(c_1) - \mathbf{y}_b(i) & >  \mathbf{y}_a(c_1) - \frac{1}{2} s_a - \mathbf{y}_b(i) \\
            & >  \mathbf{y}_a(c_1) - \frac{1}{2} s_a - \big(\mathbf{y}_a(i) + \frac{1}{2} s_a\big) \\
            & = \mathbf{y}_a(c_1) - \mathbf{y}_a(i) - s_a \geq 0
        \end{aligned}
    \end{equation}
    Therefore, $\mathbf{y}_b(c_1)$ is the largest logit in $\mathbf{y}_b$. The predicted class given by $\mathbf{y}_b$ will be the same as that of $\mathbf{y}_a$, which completes the proof.
\end{proof}

\subsection{Proof of Corollary~\ref{theorem:2}}

\label{app:proof2}
\begin{proof}
    According to the first assumption in Corollary~\ref{theorem:2}, we can have:
    \begin{equation}
        \|f_E(\mathcal{G}_w^a; \hat{\theta}) - f_E(\mathcal{G}_w^b; \hat{\theta}) \|_2 
        \leq 
        \sup_{\|\delta \| \leq \epsilon} \|f_E(\mathcal{G}_w^a; {\theta}+\delta) - f_E(\mathcal{G}_w^b; {\theta}+\delta) \|_2
    \end{equation}
    With the second assumption in Corollary~\ref{theorem:2}, we can have:
    \begin{equation}
        \frac{1}{2} \cdot \frac{\underline{s_a}}{\prod_{i=1}^K \|\mathbf{W}_i\|_2} 
        \leq 
        \frac{1}{2} \cdot \frac{s_a}{\prod_{i=1}^K \|\mathbf{W}_i\|_2}
    \end{equation}
    With the condition specified in Corollary~\ref{theorem:2}, i.e., $\sup_{\|\delta \| \leq \epsilon} \|f_E(\mathcal{G}_w^a; {\theta}+\delta) - f_E(\mathcal{G}_w^b; {\theta}+\delta) \|_2 < \frac{1}{2} \cdot \frac{\underline{s_a}}{\prod_{i=1}^K \|\mathbf{W}_i\|_2} $, we can derive that:
    \begin{equation}
        \|f_E(\mathcal{G}_w^a; \hat{\theta}) - f_E(\mathcal{G}_w^b; \hat{\theta}) \|_2 
        < 
        \frac{1}{2} \cdot \frac{s_a}{\prod_{i=1}^K \|\mathbf{W}_i\|_2}.
    \end{equation}
    According to Theorem~\ref{theorem:1}, the predicted class on the paired watermark graph $\mathcal{G}_w^b$ from $f_C$ is guaranteed to the same as that of $\mathcal{G}_w^a$, which completes the proof.
\end{proof}

    

\begin{table}[t]
    \centering
    \caption{Dataset statistics for semi-supervised learning}
    \vspace{-1em}
    \small
    \begin{tabular}{lcccc}
    \toprule 
    Dataset & \#Graphs & \#Avg. Nodes & \#Avg. Edges & \#Classes \\
    \midrule
    PROTEINS & 1,113 & 39.1 & 145.6 & 2 \\
    NCI1 & 4,110 & 29.9 & 32.3 & 2 \\
    FRANKENSTEIN & 4,337 & 16.9 & 17.9 & 2 \\
    \bottomrule
    \end{tabular}
    \label{tab:dataset_appendix_semi}
    \vspace{1em}
    \caption{Dataset statistics for transfer learning}
    \vspace{-1em}
    \small
    \begin{tabular}{llccc}
    \toprule 
    Utilization & Dataset & \#Graphs & \#Avg. Nodes & \#Avg. Edges \\
    \midrule
    Pretraining & ZINC15 & 200,000 & 23.1 & 24.8 \\
    \midrule
    \multirow{6}{*}{Finetune} & Tox21 & 7,831 & 18.6 & 38.6 \\
    & ToxCast & 8,597 & 18.7 & 38.4 \\
    & BBBP & 2,050 & 23.9 & 51.6 \\
    & BACE & 1,513 & 34.1 & 73.7 \\
    & SIDER & 1,427 & 33.6 & 70.7 \\
    & ClinTox & 1,484 & 26.1 & 55.5 \\
    \bottomrule
    \end{tabular}
    \label{tab:dataset_appendix_transfer}
    \vspace{-1.5em}
\end{table}

\section{Experimental Setting Details}
\label{sec:addition_setting}
\subsection{Datasets and Learning Settings}
\label{appendix:dataset_statistics}

\textbf{Semi-Supervised Setting}. 
In the semi-supervised setting, GNNs are trained via pre-training \& fine-tuning. For pre-training, the epoch number is set as $1000$ and the learning rate is set as $0.0003$. During fine-tuning, the label rate is set as $50\%$. We conduct experiments across various train/test splits in the downstream tasks, repeating them 50 times to derive the average results.

\noindent \textbf{Transfer Learning Setting}. 
In the transfer learning setting~\cite{hu2020pretraining}. We utilize 200,000 unlabeled molecules sampled from the ZINC15 dataset~\cite{sterling2015zinc} for pre-training, and involve 6 public molecular graph classification benchmarks in the downstream tasks. The statistics details of these datasets are summarized in Tab.~\ref{tab:dataset_appendix_transfer}. During pre-training, the epoch number is set as $200$ and the learning rate is $0.001$. The label rate in fine-tuning is set as $50\%$. We conduct experiments across various train/test splits in the downstream tasks for $10$ times and report the average results.

\begin{table*}[t]
    \centering
    \small
    \caption{Results in watermarking edge prediction-pretrained GNNs~\cite{hu2020pretraining} under semi-supervised setting.}
    \vskip -1em
    {\begin{tabular}{lllccccccc}
    \toprule
    Scenario & {Dataset} & {Metrics (\%)} & 
    {Non-Watermarked} & WPE & SSL-WM & {Deepsign}& {RWM-Pre}& {CWM-Pre}& {PreGIP}\\
     \midrule
    \multirow{6}{*}{Fix} & \multirow{3}{*}{PROTEINS} & \multirow{1}{*}{Accuracy} & {68.0$\pm$1.1} & 66.9$\pm$1.5 & 66.3$\pm$1.5 & 66.6$\pm$1.4 & 62.3$\pm$1.2 & 62.9$\pm$0.9 & \textbf{69.1$\pm$1.1}\\
        
    & & {{IP Gap} } & -  & 5.8$\pm$1.5 & 3.7$\pm$1.9 & 6.0$\pm$4.7 & 29.5$\pm$9.7 & 8.1$\pm$2.5 & \textbf{31.0$\pm$3.5}\\
    & & {{IP ROC} } & -  & 58.6 & 71.2 & 65.5& 90.0& 58.0& \textbf{100}\\
    \cmidrule{2-10}
    & \multirow{3}{*}{NCI1}
    & \multirow{1}{*}{Accuracy} & 68.7$\pm$0.9 &  65.7$\pm$1.6 & 67.9$\pm$0.9  & 69.0$\pm$1.0 & 68.4$\pm$0.8 &  68.5$\pm$0.8 & \textbf{69.7$\pm$0.6}\\

    & & {{IP Gap} } & - & 34.8$\pm$14.6 & 5.0$\pm$4.9 & 7.5$\pm$0.1 & 12.5$\pm$1.3 & 11.0$\pm$2.3 & \textbf{43.0$\pm$9.8}\\
    & & {{IP ROC} } & - & 98.4 & 80.0 & 84.0 & 78.0 & 82.5 & \textbf{100}\\
    \midrule
    \multirow{6}{*}{Finetune} & \multirow{3}{*}{PROTEINS} & \multirow{1}{*}{{Accuracy} } &  68.2$\pm$1.5 & 67.8$\pm$1.7 & 67.8$\pm$1.9 & 68.2$\pm$2.2 & 67.5$\pm$1.6 & 63.1$\pm$3.1 & \textbf{67.9$\pm$2.2}\\
        
    & & {{IP Gap} } & - & 2.3$\pm$2.2 & 1.5$\pm$1.3 & 3.0$\pm$1.0 & 6.0$\pm$5.0 & 14.0$\pm$7.5 & \textbf{35.0$\pm$1.4}\\
    & & IP ROC & - & 52.2 & 79.0 & 54.0 & 61.0& 58.5& \textbf{87.5}\\
    \cmidrule{2-10}
    & \multirow{3}{*}{NCI1}
    & \multirow{1}{*}{Accuracy} 
    & {69.6$\pm$0.8} & 67.1$\pm$1.3 & 68.8$\pm$1.2 & 68.8$\pm$0.8 &  69.7$\pm$0.9 &  69.7$\pm$0.6 & \textbf{70.8$\pm$0.9}\\
        
    & & {{IP Gap} } & -  & 39.3$\pm$38.7 & 0.0$\pm$0.0 & 4.5$\pm$0.7 & 5.0$\pm$1.5 & 9.5$\pm$3.4 & \textbf{16.5$\pm$14.7}\\
    & & IP ROC & - & 64.0 & 50.0 & 57.0 & 60.5 & 68.0 & \textbf{85.0}\\
    \bottomrule 
        \end{tabular}}
    \label{tab:semi_EdgePred_IP_performance}

    \vspace{1em}
    \caption{Results in watermarking of edge prediction-pretrained GNNs~\cite{hu2020pretraining} in the transfer learning setting.}
    \vskip -1em
    {\begin{tabularx}{0.98\linewidth}{XX>{\centering\arraybackslash}p{0.13\linewidth}CCCCCCC}
    \toprule 
    {Dataset} & {Metrics (\%)} & 
    {Non-Watermarked} & WPE & SSL-WM & {Deepsign}& {RWM-Pre}& {CWM-Pre}& {PreGIP}\\
    \midrule
    \multirow{3}{*}{Tox21}  
    & Accuracy & 92.7$\pm$0.2 & 92.6$\pm$0.2 & 92.2$\pm$0.1 &92.8$\pm$0.1 & 92.9$\pm$0.2 & 92.8 & \textbf{92.9$\pm$0.2}\\
        
    &  {{IP Gap} } & - & 0.9$\pm$0.6 & 0.3$\pm$0.2 & 4.9$\pm$11.6 & 7.5$\pm$4.8  & 11.9$\pm$10.6 & \textbf{32.4$\pm$28.0}\\
    & {{IP ROC} } & - & 90.0 & 80.0 & 62.0 & 62.0 & 66.0 & \textbf{96.0}\\
    \midrule
    \multirow{3}{*}{ToxCast}
    & Accuracy & 83.5$\pm$0.3 & 83.5$\pm$0.5 & 82.8$\pm$0.2 & 83.7$\pm$3.2 & 83.4$\pm$0.4 & 83.3$\pm$0.1 & \textbf{83.8$\pm$0.1}\\
        
    & {{IP Gap} } & - & 2.1$\pm$0.3 & 2.9$\pm$2.4 & 4.4$\pm$3.2 & 27.6$\pm$24.6  & 7.2$\pm$2.4 & \textbf{33.6$\pm$12.1}\\
    & {{IP ROC} } & - & 60.0 & 70.0 & 96.0 & 72.0 & 70.0 & \textbf{100.0}\\
    \midrule
    \multirow{3}{*}{BBBP}
    & Accuracy & 83.9$\pm$1.8 & 82.1$\pm$1.3 & 79.2$\pm$1.0 & \textbf{85.5$\pm$1.1} & 85.3$\pm$1.2 & 85.2$\pm$1.1 & {85.4$\pm$1.4}\\
        
    & {{IP Gap} } & - & 2.5$\pm$1.5 & 7.0$\pm$3.5 & 12.0$\pm$6.8 & 25.4$\pm$22.0  & 26.1$\pm$12.0 & \textbf{39.0$\pm$30.5}\\
    & {{IP ROC} } & - & 70.0 & 60.0 & 90.0 & 80.0 & 84.0 & \textbf{94.0}\\
    \midrule
    \multirow{3}{*}{BACE}
    & Accuracy & 70.2$\pm$0.8 & 68.4$\pm$0.8  & 68.9$\pm$0.5 & 73.4$\pm$2.2 & 74.3$\pm$1.8 & 74.4$\pm$1.6 & \textbf{74.5$\pm$2.4}\\
        
    & {{IP Gap} } & - & 5.3$\pm$1.5 & 17.5$\pm$12.0 & 6.7$\pm$6.4 & 26.7$\pm$25.9  & 12.2$\pm$10.4 & \textbf{38.0$\pm$26.4}\\
    & {{IP ROC} } & - & 50.0 & 80.0 & 68.0 & 80.0 & 78.0 & \textbf{84.0}\\
    \midrule
    \multirow{3}{*}{SIDER}
    & Accuracy & 75.3$\pm$0.3 & 75.2$\pm$0.5 & \textbf{75.2$\pm$0.3} &{75.1$\pm$0.3} & 74.8$\pm$0.4 & 74.7$\pm$0.3 & {74.8$\pm$0.2}\\
        
    & {{IP Gap} } & - & 2.1$\pm$1.6 & 0.8$\pm$0.6 &3.7$\pm$3.4 & 27.8$\pm$27.1  & 21.4$\pm$17.5 & \textbf{31.2$\pm$29.6}\\
    & {{IP ROC} } & - & 70.0 & 60.0 &60.0 & 80.0 & 64.0 & \textbf{80.0}\\
    \midrule
    \multirow{3}{*}{ClinTox}
    & Accuracy & 91.3$\pm$0.5 & 92.2$\pm$1.2  & 91.4$\pm$1.3  & 91.6$\pm$1.2 & 91.6$\pm$1.7 & 91.4$\pm$1.3 & \textbf{91.8$\pm$1.2}\\
        
    & {{IP Gap} } & - & 2.0$\pm$1.0 & 0.0$\pm$0.0 &2.0$\pm$1.0 & 37.2$\pm$34.0  & 33.2$\pm$31.5 & \textbf{45.0$\pm$42.0}\\
    & {{IP ROC} } & - & 80.0 & 50.0 &60.0 & 82.0 & 80.0 & \textbf{86.0}\\
    \bottomrule 
    \end{tabularx}}
    \vskip -1.em
    \label{tab:transfer_EdgeMask_IP_performance}
\end{table*}

\subsection{Details of Baselines}
\label{appendix:baselines}
\begin{itemize}[leftmargin=*]
    \item  \textbf{WPE}~\cite{wu2022watermarking}: It enforce the encoder to have dissimilar embeddings for a trigger-attached sample with the original sample. Then, the fraction of trigger-attached samples and original samples receiving the different predictions is used as the IP indication score. To extend this approach to graph-structured data, we use randomly generated graphs as triggers in the implementation. 
    \item  \textbf{SSL-WM}~\cite{lv2022ssl}: This is a black-box watermarking approach for  self-supervised learning. SSL-WM enforces the encoder to give similar embeddings to the watermark samples, which will lead to abnormal entropy of the posteriors of watermark samples. During the verification phase, fraction of abnormal posteriors of watermark samples are used as IP indication score.  To extend this method to graph-structured data, we adopt randomly generated graphs as the watermark samples. 
    \item \textbf{Deepsign}~\cite{rouhani2018deepsigns}: Deepsign enforces watermark samples to have similar embeddings, resulting in identical downstream predictions. Originally requiring intermediate representations, it is adapted here for black-box verification by measuring the fraction of watermark samples predicted as the same class.
    \item \textbf{RWM-Pre}~\cite{zhao2021watermarking}:  This baseline, inspired by RWM~\cite{zhao2021watermarking}, prompts models to classify random Erdos-Renyi graphs with random features into specific classes. Originally designed for semi-supervised node classification, we adapt it for GNN pretraining (RWM-Pre) by clustering pretrained GIN embeddings to generate pseudo classes. IP verification uses the fraction of watermark graphs predicted into the same class.
    \item \textbf{CWM-Pre}~\cite{bansal2022certifiedwm}: A certifiable watermarking method using noise injection into model parameters to predict special samples as specific classes. We adapt it for GNN pretraining by integrating its noise injection technique into RWM-Pre, naming the approach CWM-Pre.
    \item \textbf{GCBA}~\cite{zhang2023graph}: It is a backdoor attack method for graph contrastive learning. We follow the setting in~\cite{adi2018turning} to quantify the difference between the attack success rate of trigger-attached graphs.
\end{itemize}

\section{Flexibility of Watermarking}
\label{appendix:additional_results_flexibility}
The complete results of watermarking the edge prediction pretraining GNN encoder under semi-supervised learning and transfer learning settings are reported in Tab.~\ref{tab:semi_EdgePred_IP_performance} and Tab.~\ref{tab:transfer_EdgeMask_IP_performance}, respectively. The observations are rather similar to those of Fig.~\ref{fig:flexibility_of_watermarking} and Tab.~\ref{tab:transfer_GraphCL_IP_performance} in Sec.~\ref{sec:results_of_watermarking_pretraining}. These results further demonstrate the effectiveness of our {\method} in watermarking different pretraining strategies for various learning settings.

\end{document}